\documentclass[fleqn,twoside]{article}
\usepackage{espcrc1}

\newcommand{\ttbs}{\char'134}
\newcommand{\AmS}{{\protect\the\textfont2
  A\kern-.1667em\lower.5ex\hbox{M}\kern-.125emS}}

\title{Note on Evaluation of Hierarchical Modular Systems}

\author{Mark Sh. Levin
\thanks{
 Mark Sh. Levin:~
  http://www.mslevin.iitp.ru;
 email: mslevin@acm.org
  }
  }

\begin{document}

\maketitle

\begin{abstract}
 This survey note describes a brief systemic view to approaches for
 evaluation of hierarchical composite (modular) systems.
 The list of considered
 issues involves the following:
 (i) basic assessment scales
 (quantitative scale, ordinal scale, multicriteria description,
 two kinds of poset-like scales),
 (ii) basic types of scale transformations problems,
%
%
 (iii) basic types of scale integration methods.
 Evaluation of the modular systems is considered as
 assessment of system components
 (and their compatibility) and integration of the obtained
 local estimates into the total system estimate(s).
 This process is based on the above-mentioned problems
 (i.e., scale transformation and integration).
%
 Illustrations of the assessment problems and evaluation approaches
 are presented (including numerical examples).

~~

{\it Keywords:}~
                   modular systems,
                   system design,
                   system evaluation,
                   multicriteria analysis,
                   quantitative scales,
                   ordinal scales,
                   poset-like scales,
                   interval multiset estimates,
                   frameworks,
                   heuristics

\vspace{1pc}
\end{abstract}



\newcounter{cms}
\setlength{\unitlength}{1mm}

\section{Introduction}

 In recent decades,
 the significance of modular (multi-component) systems has been increased
 (e.g., \cite{bald00,clune12,hua98,lev06,lev12morph,lev13intro}).
 This survey note describes a brief systemic view to approaches for
 evaluation of hierarchical composite (modular) systems.
 The list of considered
 issues involves the following:
 ~(i) basic assessment scales
 (quantitative scale, ordinal scale, multicriteria description,
 two kinds of poset-like scales),
 ~(ii) basic types of scale transformation problems
 (i.e., mapping \(1\):
  ~{\it initial scale} \(\Rightarrow\) {\it resultant scale}),
%
%
 ~(iii) basic types of scale integration approaches.
 (i.e., mapping \(2\):~
 {\it initial scales} \(\Rightarrow\) {\it resultant integrated scale}).
%
 It is assumed that the above-mentioned mappings are
 monotone (or anti-monotone).
%
%
 Here, {\it data envelopment analysis} is not considered
 (e.g., \cite{than01}).
 Our evaluation of composite (modular) systems is examined as
 assessment of system components
 (and their compatibility) and integration of the obtained
 local estimates into the total system estimate(s)
 (e.g., \cite{lev98,levf01,lev06,lev12a}).
 Mainly, integration of component estimates
 is considered
 (estimates of system component compatibility can be examined as additional system components).
 Thus,
  the described system evaluation approach considered as a combination of the above-mentioned problems
 (i.e., transformation of scales and integration of scales).
%
%
%
 Now,
 it is reasonable to point out the following:

 (a) composite (modular) system
 (e.g., two-layer hierarchy)
 ~\(S = S_{1} \star ... \star S_{i} \star  ... \star
 S_{m}\)~
 (where
  ~\( S_{1}\), ... \(S_{i}\),  ...  \(S_{m}\)
  are the system components/parts)
 (Fig. 1)
 (e.g., \cite{lev98,lev06,lev11agg,lev12morph,lev12a}),

 (b) local domains (e.g., scales, sets of estimates) to evaluate the quality
 (excellence, ``utility'')
 of the system components
  ~\(\{ S_{1}, ... , S_{i},  ... , S_{m} \}\)
  (and/or their design alternatives DAs:  \(\{X_{i,1},...,X_{i,q_{i}} |~ i=\overline{1,m}\}\))
 and a total domain (scale, set of estimates) to evaluate the whole system ~\(S \) (Fig. 2)
 (e.g., \cite{lev98,levf01,lev06,lev12morph,lev12a}).

 Generally,
 two basic situations can be examined
 (e.g., \cite{lev98,levf01,lev06}):

  {\bf Situation 1.}
  Evaluation of a whole system
  to direct obtaining the total
  system estimate
  (e.g., expert judgment procedures,
  system testing procedures, statistical data processing,
  collection and processing of data from databases,
  technical measurement procedures, hybrid procedures).

  {\bf Situation 2.} Two-stage framework:
  2.1. Evaluation (assessment)  of system components.
  2.2. Integration of the components estimates into the total system
  estimate (this stage can be executed several times hierarchically).

 Usually, the following basic approaches are used:

 1. Expert judgment (e.g., domain experts).

 2. Measurement procedures:
 (a) technical measurement (i.e., physical system testing),
 (b) statistical measurement and data processing,
 (c) expert judgment,
%
%
 (d) assessment based on  data bases,
 and
 (e) composite (hybrid) procedure.

 3. Computer simulation.


\begin{center}
\begin{picture}(50,32)
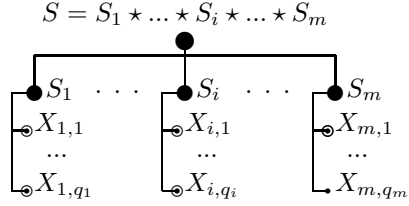
\put(00,00){\makebox(0,0)[bl]{Fig. 1. Composite (modular) system}}

\put(04,28){\makebox(0,8)[bl]{\(S = S_{1} \star ... \star S_{i}
 \star ... \star S_{m}\)}}

\put(3,24){\line(1,0){40}} \put(23,22){\line(0,1){04}}
\put(23,26){\circle*{2.6}}

\put(3,24){\line(0,-1){04}} \put(23,24){\line(0,-1){04}}
\put(43,24){\line(0,-1){04}}

\put(4.5,18){\makebox(0,8)[bl]{\(S_{1}\)}}
\put(24.5,18){\makebox(0,8)[bl]{\(S_{i}\)}}
\put(44.5,18){\makebox(0,8)[bl]{\(S_{m}\)}}

\put(3,19){\circle*{2}} \put(23,19){\circle*{2}}
\put(43,19){\circle*{2}}


\put(11,19){\makebox(0,8)[bl]{. . .}}
\put(31,19){\makebox(0,8)[bl]{. . .}}


\put(0,19){\line(0,-1){13}} \put(0,19){\line(1,0){02}}

\put(03,13){\makebox(0,8)[bl]{\(X_{1,1}\)}}


\put(0,14){\line(1,0){02}}

\put(2,14){\circle*{0.8}} \put(2,14){\circle{1.5}}


\put(04.5,10.5){\makebox(0,8)[bl]{...}}


\put(03,05){\makebox(0,8)[bl]{\(X_{1,q_{1}}\)}}

\put(0,06){\line(1,0){02}}

\put(2,06){\circle*{0.8}} \put(2,06){\circle{1.5}}


\put(20,19){\line(0,-1){13}} \put(20,19){\line(1,0){02}}

\put(23,13){\makebox(0,8)[bl]{\(X_{i,1}\)}}

\put(20,14){\line(1,0){02}}

\put(22,14){\circle*{0.8}} \put(22,14){\circle{1.5}}


\put(24.5,10.5){\makebox(0,8)[bl]{...}}


\put(23,05){\makebox(0,8)[bl]{\(X_{i,q_{i}}\)}}

\put(20,06){\line(1,0){02}}

\put(22,06){\circle*{0.8}} \put(22,06){\circle{1.5}}


\put(40,19){\line(0,-1){13}} \put(40,19){\line(1,0){02}}

\put(43,13){\makebox(0,8)[bl]{\(X_{m,1}\)}}

\put(40,14){\line(1,0){02}}

\put(42,14){\circle*{0.8}} \put(42,14){\circle{1.5}}


\put(44.5,10.5){\makebox(0,8)[bl]{...}}


\put(43,05){\makebox(0,8)[bl]{\(X_{m,q_{m}}\)}}

\put(40,06){\line(1,0){02}}

\put(42,06){\circle*{0.8}} \put(42,14){\circle{1.5}}

\end{picture}
\end{center}

 In this paper, the following evaluation problems are examined:
%
 ~(1) assessment of DAs for leaf nodes  of the system model
 (i.e., system components)
 (e.g., quantitative scale, ordinal scale, multicriteria
 description, poset-like scales);
 ~(2) integration of the obtained estimates for DAs
 to obtain
 the integrated (total) estimate for the composite final system
 (or its versions).
 An illustration of the evaluation procedure for two-layer system is presented in
 Fig. 2.

\begin{center}
\begin{picture}(80,62)
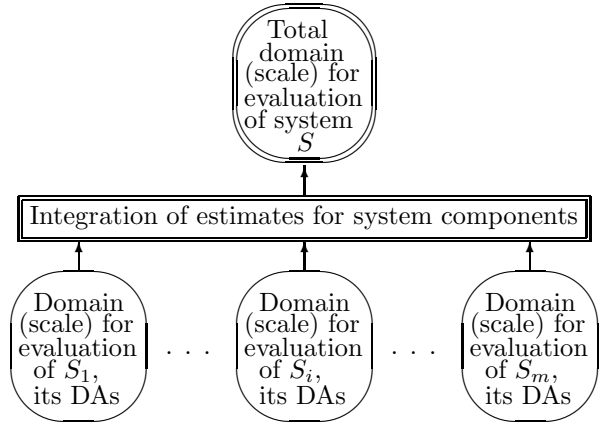
\put(02.2,00){\makebox(0,0)[bl]{Fig. 2. Evaluation scheme for
 two-layer system}}


\put(09,25){\vector(0,1){4}}

\put(09,15){\oval(18,20)}

\put(03,19.7){\makebox(0,0)[bl]{Domain }}
\put(01,16.1){\makebox(0,0)[bl]{(scale) for}}
\put(01,13.7){\makebox(0,0)[bl]{evaluation}}
\put(03,10.1){\makebox(0,0)[bl]{of \(S_{1}\),}}
\put(03,07.3){\makebox(0,0)[bl]{its DAs}}


\put(20.5,14){\makebox(0,8)[bl]{. . .}}


\put(39,25){\vector(0,1){04}}

\put(39,15){\oval(18,20)}

\put(33,19.7){\makebox(0,0)[bl]{Domain }}
\put(31,16.1){\makebox(0,0)[bl]{(scale) for}}
\put(31,13.7){\makebox(0,0)[bl]{evaluation}}
\put(33,10.1){\makebox(0,0)[bl]{of \(S_{i}\), }}
\put(33,07.3){\makebox(0,0)[bl]{its DAs}}


\put(50.5,14){\makebox(0,8)[bl]{. . .}}


\put(69,25){\vector(0,1){04}}

\put(69,15){\oval(18,20)}

\put(63,19.7){\makebox(0,0)[bl]{Domain }}
\put(61,16.1){\makebox(0,0)[bl]{(scale) for}}
\put(61,13.7){\makebox(0,0)[bl]{evaluation}}
\put(63,10.1){\makebox(0,0)[bl]{of \(S_{m}\), }}
\put(63,07.3){\makebox(0,0)[bl]{its DAs}}


\put(01,29){\line(1,0){76}} \put(01,35){\line(1,0){76}}
\put(01,29){\line(0,1){06}} \put(77,29){\line(0,1){06}}

\put(01.5,29.4){\line(1,0){75}} \put(01.5,34.5){\line(1,0){75}}
\put(01.5,29.4){\line(0,1){05}} \put(76.5,29.5){\line(0,1){05}}

\put(02.5,30.6){\makebox(0,0)[bl]{Integration of estimates for
 system components}}

\put(39,35){\vector(0,1){4}}


\put(39,50){\oval(18,20)} \put(39,50){\oval(19,21)}

\put(34,56){\makebox(0,0)[bl]{Total}}
\put(33,53){\makebox(0,0)[bl]{domain }}
\put(31,49.4){\makebox(0,0)[bl]{(scale) for}}
\put(31,47){\makebox(0,0)[bl]{evaluation}}
\put(31,43.4){\makebox(0,0)[bl]{of system  }}
\put(38,41){\makebox(0,0)[bl]{\(S\)}}




\end{picture}
\end{center}

 An example of three-layer system structure is presented
 in Fig. 3.

\begin{center}
\begin{picture}(39.5,55)
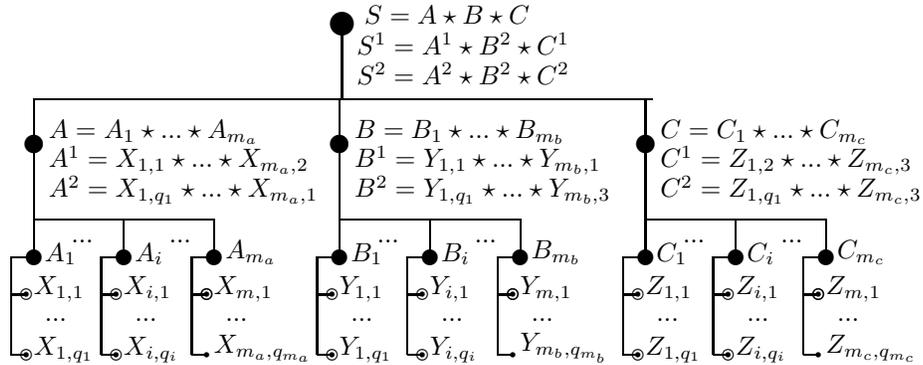
\put(20,00){\makebox(0,0)[bl]{Fig. 3. Three-layer composite
 (modular) system}}

\put(47,50){\makebox(0,8)[bl]{\(S = A \star B  \star C\)}}

\put(46,46){\makebox(0,8)[bl]{\(S^{1} = A^{1} \star B^{2}  \star
C^{1}\)}}

\put(46,42){\makebox(0,8)[bl]{\(S^{2} = A^{2} \star B^{2}  \star
C^{2}\)}}

\put(44,40){\line(0,1){10}} \put(44,50){\circle*{3}}


\put(3,40){\line(1,0){82.2}}


\put(05,34){\makebox(0,8)[bl]{\(A = A_{1} \star ...
 \star A_{m_{a}}\)}}

 \put(05,30){\makebox(0,8)[bl]{\(A^{1} = X_{1,1} \star ...
 \star X_{m_{a},2}\)}}

\put(05,26){\makebox(0,8)[bl]{\(A^{2} = X_{1,q_{1}} \star ...
 \star X_{m_{a},1}\)}}

\put(3,24){\line(0,1){16}} \put(3,34){\circle*{2.4}}

\put(3,24){\line(1,0){24}}

\put(3,24){\line(0,-1){04}} \put(15,24){\line(0,-1){04}}
\put(27,24){\line(0,-1){04}}

\put(4.5,18){\makebox(0,8)[bl]{\(A_{1}\)}}
\put(16.5,18){\makebox(0,8)[bl]{\(A_{i}\)}}
\put(28.5,18){\makebox(0,8)[bl]{\(A_{m_{a}}\)}}

\put(3,19){\circle*{2}} \put(15,19){\circle*{2}}
\put(27,19){\circle*{2}}


\put(08,21){\makebox(0,8)[bl]{...}}
\put(21,21){\makebox(0,8)[bl]{...}}


\put(0,19){\line(0,-1){13}} \put(0,19){\line(1,0){02}}

\put(03,13){\makebox(0,8)[bl]{\(X_{1,1}\)}}


\put(0,14){\line(1,0){02}}

\put(2,14){\circle*{0.8}} \put(2,14){\circle{1.5}}


\put(04.5,10.5){\makebox(0,8)[bl]{...}}


\put(03,05){\makebox(0,8)[bl]{\(X_{1,q_{1}}\)}}

\put(0,06){\line(1,0){02}}

\put(2,06){\circle*{0.8}} \put(2,06){\circle{1.5}}


\put(12,19){\line(0,-1){13}} \put(12,19){\line(1,0){02}}

\put(15,13){\makebox(0,8)[bl]{\(X_{i,1}\)}}

\put(12,14){\line(1,0){02}}

\put(14,14){\circle*{0.8}} \put(14,14){\circle{1.5}}


\put(16.5,10.5){\makebox(0,8)[bl]{...}}


\put(15,05){\makebox(0,8)[bl]{\(X_{i,q_{i}}\)}}

\put(12,06){\line(1,0){02}}

\put(14,06){\circle*{0.8}} \put(14,06){\circle{1.5}}


\put(24,19){\line(0,-1){13}} \put(24,19){\line(1,0){02}}

\put(27,13){\makebox(0,8)[bl]{\(X_{m,1}\)}}

\put(24,14){\line(1,0){02}}

\put(26,14){\circle*{0.8}} \put(26,14){\circle{1.5}}


\put(28.5,10.5){\makebox(0,8)[bl]{...}}


\put(27,05){\makebox(0,8)[bl]{\(X_{m_{a},q_{m_{a}}}\)}}

\put(24,06){\line(1,0){02}}

\put(26,06){\circle*{0.8}} \put(26,14){\circle{1.5}}

\end{picture}
%
\begin{picture}(39.5,50)

\put(05,34){\makebox(0,8)[bl]{\(B = B_{1} \star ...
 \star B_{m_{b}}\)}}

 \put(05,30){\makebox(0,8)[bl]{\(B^{1} = Y_{1,1} \star ...
 \star Y_{m_{b},1}\)}}

\put(05,26){\makebox(0,8)[bl]{\(B^{2} = Y_{1,q_{1}} \star ...
 \star Y_{m_{b},3}\)}}

\put(3,24){\line(0,1){16}} \put(3,34){\circle*{2.4}}

\put(3,24){\line(1,0){24}}

\put(3,24){\line(0,-1){04}} \put(15,24){\line(0,-1){04}}
\put(27,24){\line(0,-1){04}}

\put(4.5,18){\makebox(0,8)[bl]{\(B_{1}\)}}
\put(16.5,18){\makebox(0,8)[bl]{\(B_{i}\)}}
\put(28.5,18){\makebox(0,8)[bl]{\(B_{m_{b}}\)}}

\put(3,19){\circle*{2}} \put(15,19){\circle*{2}}
\put(27,19){\circle*{2}}


\put(08,21){\makebox(0,8)[bl]{...}}
\put(21,21){\makebox(0,8)[bl]{...}}


\put(0,19){\line(0,-1){13}} \put(0,19){\line(1,0){02}}

\put(03,13){\makebox(0,8)[bl]{\(Y_{1,1}\)}}


\put(0,14){\line(1,0){02}}

\put(2,14){\circle*{0.8}} \put(2,14){\circle{1.5}}


\put(04.5,10.5){\makebox(0,8)[bl]{...}}


\put(03,05){\makebox(0,8)[bl]{\(Y_{1,q_{1}}\)}}

\put(0,06){\line(1,0){02}}

\put(2,06){\circle*{0.8}} \put(2,06){\circle{1.5}}


\put(12,19){\line(0,-1){13}} \put(12,19){\line(1,0){02}}

\put(15,13){\makebox(0,8)[bl]{\(Y_{i,1}\)}}

\put(12,14){\line(1,0){02}}

\put(14,14){\circle*{0.8}} \put(14,14){\circle{1.5}}


\put(16.5,10.5){\makebox(0,8)[bl]{...}}


\put(15,05){\makebox(0,8)[bl]{\(Y_{i,q_{i}}\)}}

\put(12,06){\line(1,0){02}}

\put(14,06){\circle*{0.8}} \put(14,06){\circle{1.5}}


\put(24,19){\line(0,-1){13}} \put(24,19){\line(1,0){02}}

\put(27,13){\makebox(0,8)[bl]{\(Y_{m,1}\)}}

\put(24,14){\line(1,0){02}}

\put(26,14){\circle*{0.8}} \put(26,14){\circle{1.5}}


\put(28.5,10.5){\makebox(0,8)[bl]{...}}


\put(27,05){\makebox(0,8)[bl]{\(Y_{m_{b},q_{m_{b}}}\)}}

\put(24,06){\line(1,0){02}}

\put(26,06){\circle*{0.8}} \put(26,14){\circle{1.5}}

\end{picture}
%
\begin{picture}(38,50)



\put(05,34){\makebox(0,8)[bl]{\(C = C_{1} \star ...
 \star C_{m_{c}}\)}}

 \put(05,30){\makebox(0,8)[bl]{\(C^{1} = Z_{1,2} \star ...
 \star Z_{m_{c},3}\)}}

\put(05,26){\makebox(0,8)[bl]{\(C^{2} = Z_{1,q_{1}} \star ...
 \star Z_{m_{c},3}\)}}

\put(3,24){\line(0,1){16}} \put(3,34){\circle*{2.4}}

\put(3,24){\line(1,0){24}}

\put(3,24){\line(0,-1){04}} \put(15,24){\line(0,-1){04}}
\put(27,24){\line(0,-1){04}}

\put(4.5,18){\makebox(0,8)[bl]{\(C_{1}\)}}
\put(16.5,18){\makebox(0,8)[bl]{\(C_{i}\)}}
\put(28.5,18){\makebox(0,8)[bl]{\(C_{m_{c}}\)}}

\put(3,19){\circle*{2}} \put(15,19){\circle*{2}}
\put(27,19){\circle*{2}}


\put(08,21){\makebox(0,8)[bl]{...}}
\put(21,21){\makebox(0,8)[bl]{...}}


\put(0,19){\line(0,-1){13}} \put(0,19){\line(1,0){02}}

\put(03,13){\makebox(0,8)[bl]{\(Z_{1,1}\)}}


\put(0,14){\line(1,0){02}}

\put(2,14){\circle*{0.8}} \put(2,14){\circle{1.5}}


\put(04.5,10.5){\makebox(0,8)[bl]{...}}


\put(03,05){\makebox(0,8)[bl]{\(Z_{1,q_{1}}\)}}

\put(0,06){\line(1,0){02}}

\put(2,06){\circle*{0.8}} \put(2,06){\circle{1.5}}


\put(12,19){\line(0,-1){13}} \put(12,19){\line(1,0){02}}

\put(15,13){\makebox(0,8)[bl]{\(Z_{i,1}\)}}

\put(12,14){\line(1,0){02}}

\put(14,14){\circle*{0.8}} \put(14,14){\circle{1.5}}


\put(16.5,10.5){\makebox(0,8)[bl]{...}}


\put(15,05){\makebox(0,8)[bl]{\(Z_{i,q_{i}}\)}}

\put(12,06){\line(1,0){02}}

\put(14,06){\circle*{0.8}} \put(14,06){\circle{1.5}}


\put(24,19){\line(0,-1){13}} \put(24,19){\line(1,0){02}}

\put(27,13){\makebox(0,8)[bl]{\(Z_{m,1}\)}}

\put(24,14){\line(1,0){02}}

\put(26,14){\circle*{0.8}} \put(26,14){\circle{1.5}}


\put(28.5,10.5){\makebox(0,8)[bl]{...}}


\put(27,05){\makebox(0,8)[bl]{\(Z_{m_{c},q_{m_{c}}}\)}}

\put(24,06){\line(1,0){02}}

\put(26,06){\circle*{0.8}} \put(26,14){\circle{1.5}}

\end{picture}
\end{center}

 Here, the following evaluation problems are considered:
 ~(1) assessment of DAs for leaf nodes  of the system model
 (i.e., system components)
 (e.g., quantitative scale, ordinal scale, multicriteria
 description, poset-like scales);
 ~(2) integration of the obtained estimates for DAs to obtain
 integrated estimates for the composite system
 nodes (i.e., system parts, at the higher system hierarchy);
 ~(3) integration of the obtained estimates
 for system parts to obtain
 integrated (total) estimates for the final system
 (or its versions).
 An illustration of the evaluation procedure for three-layer system is presented in
 Fig. 4.

\begin{center}
\begin{picture}(117,97)

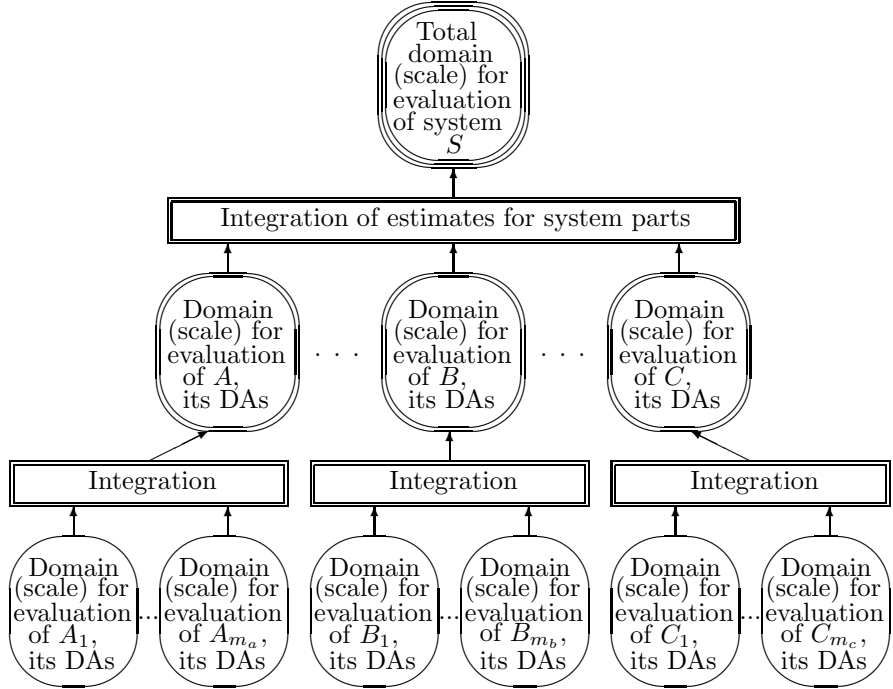
\put(20.5,00){\makebox(0,0)[bl]{Fig. 4. Evaluation scheme for
 three-layer system}}


\put(08.5,25){\vector(0,1){4}}

\put(08.5,15){\oval(17,20)}

\put(02.5,19.7){\makebox(0,0)[bl]{Domain }}
\put(0.5,16.1){\makebox(0,0)[bl]{(scale) for}}
\put(0.5,13.7){\makebox(0,0)[bl]{evaluation}}
\put(02.5,10.1){\makebox(0,0)[bl]{of \(A_{1}\),}}
\put(02.5,07.3){\makebox(0,0)[bl]{its DAs}}


\put(17,14){\makebox(0,8)[bl]{...}}


\put(29,25){\vector(0,1){04}}

\put(28.5,15){\oval(17,20)}

\put(22.5,19.7){\makebox(0,0)[bl]{Domain }}
\put(20.5,16.1){\makebox(0,0)[bl]{(scale) for}}
\put(20.5,13.7){\makebox(0,0)[bl]{evaluation}}
\put(22.5,10.1){\makebox(0,0)[bl]{of \(A_{m_{a}}\),}}
\put(22.5,07.3){\makebox(0,0)[bl]{its DAs}}


\put(00,29){\line(1,0){37}} \put(00,35){\line(1,0){37}}
\put(00,29){\line(0,1){06}} \put(37,29){\line(0,1){06}}

\put(00.5,29.4){\line(1,0){36}} \put(00.5,34.5){\line(1,0){36}}
\put(00.5,29.4){\line(0,1){05}} \put(36.5,29.5){\line(0,1){05}}

\put(10.5,30.6){\makebox(0,0)[bl]{Integration}}

\put(18.5,35){\vector(2,1){8}}


\put(48.5,25){\vector(0,1){4}}

\put(48.5,15){\oval(17,20)}

\put(42.5,19.7){\makebox(0,0)[bl]{Domain }}
\put(40.5,16.1){\makebox(0,0)[bl]{(scale) for}}
\put(40.5,13.7){\makebox(0,0)[bl]{evaluation}}
\put(42.5,10.1){\makebox(0,0)[bl]{of \(B_{1}\),}}
\put(42.5,07.3){\makebox(0,0)[bl]{its DAs}}


\put(57,14){\makebox(0,8)[bl]{...}}


\put(69,25){\vector(0,1){04}}

\put(68.5,15){\oval(17,20)}

\put(62.5,19.7){\makebox(0,0)[bl]{Domain }}
\put(60.5,16.1){\makebox(0,0)[bl]{(scale) for}}
\put(60.5,13.7){\makebox(0,0)[bl]{evaluation}}
\put(62.5,10.1){\makebox(0,0)[bl]{of \(B_{m_{b}}\),}}
\put(62.5,07.3){\makebox(0,0)[bl]{its DAs}}


\put(40,29){\line(1,0){37}} \put(40,35){\line(1,0){37}}
\put(40,29){\line(0,1){06}} \put(77,29){\line(0,1){06}}

\put(40.5,29.4){\line(1,0){36}} \put(40.5,34.5){\line(1,0){36}}
\put(40.5,29.4){\line(0,1){05}} \put(76.5,29.5){\line(0,1){05}}

\put(50.5,30.6){\makebox(0,0)[bl]{Integration}}

\put(58.5,35){\vector(0,1){4}}


\put(88.5,25){\vector(0,1){04}}

\put(88.5,15){\oval(17,20)}

\put(82.5,19.7){\makebox(0,0)[bl]{Domain }}
\put(80.5,16.1){\makebox(0,0)[bl]{(scale) for}}
\put(80.5,13.7){\makebox(0,0)[bl]{evaluation}}
\put(82.5,10.1){\makebox(0,0)[bl]{of \(C_{1}\), }}
\put(82.5,07.3){\makebox(0,0)[bl]{its DAs}}


\put(97,14){\makebox(0,8)[bl]{...}}


\put(109,25){\vector(0,1){04}}

\put(108.5,15){\oval(17,20)}

\put(102.5,19.7){\makebox(0,0)[bl]{Domain }}
\put(100.5,16.1){\makebox(0,0)[bl]{(scale) for}}
\put(100.5,13.7){\makebox(0,0)[bl]{evaluation}}
\put(102.5,10.1){\makebox(0,0)[bl]{of \(C_{m_{c}}\), }}
\put(102.5,07.3){\makebox(0,0)[bl]{its DAs}}


\put(80,29){\line(1,0){37}} \put(80,35){\line(1,0){37}}
\put(80,29){\line(0,1){06}} \put(117,29){\line(0,1){06}}

\put(80.5,29.4){\line(1,0){36}} \put(80.5,34.5){\line(1,0){36}}
\put(80.5,29.4){\line(0,1){05}} \put(116.5,29.5){\line(0,1){05}}

\put(90.5,30.6){\makebox(0,0)[bl]{Integration}}

\put(98.5,35){\vector(-2,1){8}}



\put(29,60){\vector(0,1){4}}

\put(29,49.5){\oval(18,20)} \put(29,49.5){\oval(19,21)}

\put(23,54){\makebox(0,0)[bl]{Domain }}
\put(21,50.4){\makebox(0,0)[bl]{(scale) for}}
\put(21,48){\makebox(0,0)[bl]{evaluation}}
\put(23,44.4){\makebox(0,0)[bl]{of \(A\), }}
\put(23,42){\makebox(0,0)[bl]{its DAs}}


\put(40.5,49){\makebox(0,8)[bl]{. . .}}


\put(59,60){\vector(0,1){04}}

\put(59,49.5){\oval(18,20)} \put(59,49.5){\oval(19,21)}

\put(53,54){\makebox(0,0)[bl]{Domain }}
\put(51,50.4){\makebox(0,0)[bl]{(scale) for}}
\put(51,48){\makebox(0,0)[bl]{evaluation}}
\put(53,44.4){\makebox(0,0)[bl]{of \(B\), }}
\put(53,42){\makebox(0,0)[bl]{its DAs}}


\put(70.5,49){\makebox(0,8)[bl]{. . .}}


\put(89,60){\vector(0,1){04}}

\put(89,49.5){\oval(18,20)} \put(89,49.5){\oval(19,21)}

\put(83,54){\makebox(0,0)[bl]{Domain }}
\put(81,50.4){\makebox(0,0)[bl]{(scale) for}}
\put(81,48){\makebox(0,0)[bl]{evaluation}}
\put(83,44.4){\makebox(0,0)[bl]{of \(C\), }}
\put(83,42){\makebox(0,0)[bl]{its DAs}}


\put(21,64){\line(1,0){76}} \put(21,70){\line(1,0){76}}
\put(21,64){\line(0,1){06}} \put(97,64){\line(0,1){06}}

\put(21.5,64.4){\line(1,0){75}} \put(21.5,69.5){\line(1,0){75}}
\put(21.5,64.4){\line(0,1){05}} \put(96.5,64.5){\line(0,1){05}}

\put(28,65.6){\makebox(0,0)[bl]{Integration of estimates for
 system parts}}

\put(59,70){\vector(0,1){4}}



\put(59,85){\oval(18,20)} \put(59,85){\oval(19,21)}
\put(59,85){\oval(20,22)}

\put(54,91){\makebox(0,0)[bl]{Total}}
\put(53,88){\makebox(0,0)[bl]{domain }}
\put(51,84.4){\makebox(0,0)[bl]{(scale) for}}
\put(51,82){\makebox(0,0)[bl]{evaluation}}
\put(51,78.4){\makebox(0,0)[bl]{of system  }}
\put(58,76){\makebox(0,0)[bl]{\(S\)}}




\end{picture}
\end{center}

\section{Considered Types of Assessment Scales}

 Table 1 contains  the considered
 types of assessment scales (for system parts/components, for final
 system):
 quantitative scale, ordinal scale, multicriteria description,
 poset-like scales
 (e.g., \cite{fis70,kee76,lev98,levf01,lev06,lev12a,mirkin79,roy96,ste86,zap02}).

\begin{center}
\begin{picture}(115,43)

\put(17,39){\makebox(0,0)[bl]{Table 1. Considered types of
 system assessment scales}}

\put(00,0){\line(1,0){115}} \put(00,31){\line(1,0){115}}
\put(00,37){\line(1,0){115}}

\put(00,00){\line(0,1){37}} \put(85,00){\line(0,1){37}}
\put(115,00){\line(0,1){37}}


\put(01,32.5){\makebox(0,0)[bl]{Types of scales (descriptions)}}

\put(86,33){\makebox(0,0)[bl]{Sources}}


\put(01,25.5){\makebox(0,0)[bl]{1.Quantitative scale}}


\put(86,26){\makebox(0,0)[bl]{\cite{fis70,kee76,ste86} }}


\put(01,21){\makebox(0,0)[bl]{2.Ordinal scale}}

\put(86,21){\makebox(0,0)[bl]{\cite{brans84,larmf86,lev98,mirkin79,roy96,zap02}
}}


\put(01,15.5){\makebox(0,0)[bl]{3.Multicriteria description
(vector-like estimate}}

\put(03,11.5){\makebox(0,0)[bl]{based on quantitative and/or
ordinal estimates)}}


\put(86,16){\makebox(0,0)[bl]{\cite{kee76,mirkin79,pareto71,roy96,ste86}
}}


\put(01,07){\makebox(0,0)[bl]{4.Poset-like scale based on ordinal
 estimates}}


\put(86,07){\makebox(0,0)[bl]{\cite{lev98,levf01,lev06,lev12morph}}}


\put(01,02){\makebox(0,0)[bl]{5.Poset-like scale based on
 interval multiset estimates}}



\put(86,02){\makebox(0,0)[bl]{\cite{lev12a}}}


\end{picture}
\end{center}



 Let us consider  illustrations for the above-mentioned
 basic assessment scales.

 First,
 Fig. 5 depicts illustrations for quantitative scale,
 qualitative ordinal scale, and
 multicriteria description:

 (a) quantitative scale,
 e.g., interval \((\beta, \alpha)\),
 \(\alpha\) corresponds to the best point,
 \(\beta\) corresponds to the worst point (Fig. 5a);

 (b) qualitative (ordinal) scale:
 \([1,2,...,\kappa]\), \(1\) corresponds to the best point,
 i.e., point \(i\) dominates point \(i+1\) (Fig. 5b);
 and

 (c) multicriteria description (i.e., vector-like estimates) (Fig. 5c).

 Note,
 domination binary relations
for the points, which belong to the scales
 in cases (a) and (b), are evident.
 In the case (c),
 domination is illustrated in Fig. 5:
 \(\alpha_{2} \succ  \beta_{2}\),
 \(\alpha_{2} \succ  \beta_{3}\),
 \(\alpha_{2} \succ  \beta_{4}\).
 In the case of domination by Pareto-rule
 (e.g., \cite{mirkin79,pareto71}),
 the basic domination binary relation is extended by cases as
 \(\alpha_{2} \succ^{P}  \beta_{1}\).
 Here,
 the following ordered layers of quality
 can be considered
 (as a special  ordinal scale \(D\),
 by illustration in Fig. 5c):

 (i) the ideal point (the best point)
 \(\alpha^{I}\),

 (ii) a layer of Pareto-efficient points
 (e.g., points: ~\(\{\alpha_{1},\alpha_{2},\alpha_{3},\alpha_{4}\}\)),

 (iii) near Pareto-efficient points
 (the points are close to the Pareto-layer,
 e.g., points: ~\(\{\beta_{1},\beta_{2},\beta_{3},\beta_{4},\beta_{5}\}\)),

 (iv) a next layer of quality
 (i.e, between near Pareto-efficient points and the worst point,
 e.g., points: ~\(\{\gamma_{1},\gamma_{2}\}\)),
 and

 ~(v) the worst point.

\begin{center}
\begin{picture}(54,49)

\put(05.5,00){\makebox(0,0)[bl]{Fig. 5. Quantitative scale,
ordinal
 scale, multicriteria description}}

\put(00,9){\makebox(0,0)[bl]{(a) Quantita-}}
\put(05.7,6){\makebox(0,0)[bl]{tive scale}}

\put(08.5,18){\line(1,0){3}}

\put(08,13.5){\makebox(0,0)[bl]{(\(0\))}}

\put(10,18){\vector(0,1){26.5}}


\put(12,38.5){\makebox(0,0)[bl]{\(\alpha\)}}

\put(10,39){\circle*{1.4}} \put(10,39){\circle{2.4}}

\put(00,39){\makebox(0,0)[bl]{Best}}
\put(00,36){\makebox(0,0)[bl]{point}}

\put(12,21.5){\makebox(0,0)[bl]{\(\beta\)}}

\put(10,23){\circle*{1.5}}

\put(00,24){\makebox(0,0)[bl]{Worst}}
\put(00,21){\makebox(0,0)[bl]{point}}




\put(28,9){\makebox(0,0)[bl]{(b) Ordinal}}
\put(35,6){\makebox(0,0)[bl]{scale}}

\put(29,41){\makebox(0,0)[bl]{Best}}
\put(29,38){\makebox(0,0)[bl]{point}}

\put(40,41){\circle*{1.4}} \put(40,41){\circle{2.4}}
\put(42,40.5){\makebox(0,0)[bl]{\(1\)}}

\put(40,41){\vector(0,-1){4}}

\put(42,35.5){\makebox(0,0)[bl]{\(2\)}}

\put(40,36){\circle*{1.9}} \put(40,36){\vector(0,-1){4}}

\put(42,30.5){\makebox(0,0)[bl]{\(3\)}}

\put(40,31){\circle*{1.9}} \put(40,31){\vector(0,-1){4}}

\put(38.5,24){\makebox(0,0)[bl]{...}}


\put(40,22){\vector(0,-1){4}}


\put(42,16.5){\makebox(0,0)[bl]{\(\kappa\)}}

\put(40,17){\circle*{1.9}}

\put(29,17.5){\makebox(0,0)[bl]{Worst}}
\put(29,14.5){\makebox(0,0)[bl]{point}}

\end{picture}
%
\begin{picture}(58,49)


\put(12.5,9){\makebox(0,0)[bl]{(c) Multicriteria}}
\put(19.5,6){\makebox(0,0)[bl]{description}}

\put(04,17){\circle*{0.9}}

\put(05,20.2){\makebox(0,0)[bl]{Worst}}
\put(05,17.2){\makebox(0,0)[bl]{point}}

\put(00,12.5){\makebox(0,0)[bl]{\((0,0)\)}}

\put(04,17){\vector(0,1){25}} \put(04,17){\vector(1,0){51}}

\put(40,13){\makebox(0,0)[bl]{Criterion 2}}
\put(00,43){\makebox(0,0)[bl]{Criterion 1}}


\put(06,37){\line(1,0){4}} \put(12,37){\line(1,0){4}}
\put(18,37){\line(1,0){4}} \put(24,37){\line(1,0){4}}
\put(30,37){\line(1,0){4}} \put(36,37){\line(1,0){4}}
\put(42,37){\line(1,0){4}}



\put(49,18){\line(0,1){4}} \put(49,24){\line(0,1){4}}
\put(49,30){\line(0,1){4}}


\put(49,37){\circle*{1}} \put(49,37){\circle{2}}

\put(46,42){\makebox(0,0)[bl]{Ideal}}
\put(46,39){\makebox(0,0)[bl]{point}}
\put(51,36){\makebox(0,0)[bl]{\(\alpha^{I}\)}}


\put(24,37){\circle*{1.1}} \put(24,37){\circle{1.9}}
\put(22,39){\makebox(0,0)[bl]{\(\alpha_{1}\)}}

\put(29,34){\circle*{1.1}} \put(29,34){\circle{1.9}}
\put(30.5,33){\makebox(0,0)[bl]{\(\alpha_{2}\)}}
\put(29,34){\line(-1,0){25}} \put(29,34){\line(0,-1){17}}

\put(40,27){\circle*{1.1}} \put(40,27){\circle{1.9}}
\put(41.5,27){\makebox(0,0)[bl]{\(\alpha_{3}\)}}

\put(49,22){\circle*{1.1}} \put(49,22){\circle{1.9}}
\put(51,21){\makebox(0,0)[bl]{\(\alpha_{4}\)}}


\put(07,34){\circle*{1.4}}
\put(05,29.5){\makebox(0,0)[bl]{\(\beta_{1}\)}}

\put(13,31){\circle*{1.4}}
\put(11.5,26.5){\makebox(0,0)[bl]{\(\beta_{2}\)}}

\put(18,27){\circle*{1.4}}
\put(16.5,22.5){\makebox(0,0)[bl]{\(\beta_{3}\)}}

\put(27,23){\circle*{1.4}}
\put(24.5,18.5){\makebox(0,0)[bl]{\(\beta_{4}\)}}

\put(36,18){\circle*{1.4}}
\put(34,19){\makebox(0,0)[bl]{\(\beta_{5}\)}}


\put(04,26){\circle{0.75}} \put(04,26){\circle{1.4}}
\put(5.3,25){\makebox(0,0)[bl]{\(\gamma_{1}\)}}

\put(19,17){\circle{0.75}} \put(19,17){\circle{1.4}}
\put(19.8,17.3){\makebox(0,0)[bl]{\(\gamma_{2}\)}}

\end{picture}
\end{center}


 The description of poset-like scales (or lattices) for quality of
 composite (modular) systems
 (based on ordinal estimates of DAs and their compatibility)
%
  was suggested within framework of HMMD
 approach
 (e.g., \cite{lev98,levf01,lev06,lev12morph}).
 Here, two cases have to be examined:
 (1) scale for system quality based on system components ordinal estimates
 (\(\iota = \overline{1,l}\);
      \(1\) corresponds to the best one);
 (2) scale for system quality while taking into account
 system components ordinal estimates
 and ordinal compatibility estimates between the system components
  (\(w=\overline{1,\nu}\); \(\nu\) corresponds to the best level).
%

 For the system consisting of \(m\) parts/components,
  a discrete space (poset, lattice) of the system quality (excellence) on the basis of the
 following vector is used:
 ~\(N(S)=(w(S);n(S))\),
 ~where \(w(S)\) is the minimum of pairwise compatibility
 between DAs which correspond to different system components
 (i.e.,
 \(~\forall ~P_{j_{1}}\) and \( P_{j_{2}}\),
 \(1 \leq j_{1} \neq j_{2} \leq m\))
 in \(S\),
 ~\(n(S)=(\eta_{1},...,\eta_{r},...,\eta_{k})\),
 ~where ~\(\eta_{r}\) is the number of DAs of the \(r\)th quality in ~\(S\)
 ~(\(\sum^{k}_{r=1} n_{r} = m \)).

 An example of the  three-component system
 ~\(S = X \star Y \star Z\)
 is considered (Fig. 1).
 The following ordinal scales are used:
 (a) ordinal scale for elements (priorities) is \([1,2,3]\),
 (b) ordinal scale for compatibility is \([0,1,2,3,4]\).
 For this case,
 Fig. 6a depicts the poset of system quality by components and
 Fig. 6b depicts an integrated poset with compatibility
 (each triangle corresponds to the poset from Fig. 6a).

    Generally, the following layers of system excellence can be considered
    (Fig. 6b, this corresponds to the resultant system scale \(D\) in Fig. 5b):

  {\it 1.} The ideal point \(N(S^{I})\) (\(S^{I}\) is the ideal system solution).

  {\it 2.} A layer of Pareto-efficient solutions:
  \( \{ S_{1}^{p}, S_{2}^{p}, S_{3}^{p} \} \);
  estimates are:
 \(N(S^{p}_{1}) = (2;3,0,0)\),
 \(N(S^{p}_{2})=(3;1,1,1)\), and
 \(N(S^{p}_{3})=(4;0,2,1)\).

  {\it 3.} A next layer of quality
 (e.g., neighborhood of Pareto-efficient solutions layer):
 \( \{ S'_{1}, S'_{2}, S'_{3} \} \);
 estimates are:
  \(N(S'_{1}) = (1;3,0,0)\),
 \(N(S'_{2})=(2;1,1,1)\), and
 \(N(S'_{3})=(3;0,2,1)\);

  a composite solution of this set can be
  transformed into a Pareto-efficient solution  on the basis of a
 simple improvement action(s)
 (e.g., as modification of the only one  element).

  {\it 4.} A next layer of quality \(S''\);
  estimate is:
  \(N(S'')=(1;0,3,0)\).

 {\it 5.} The worst point
 \(S_{0}\);
 estimate is:
 \(N(S_{0}) = (1;0,0,3)\).

 Note,
 the compatibility component of vector ~\(N(S)\)
 can be considered on the basis of a poset-like scale too
 (as \(n(S)\))
  (\cite{levf01,lev06}).
 In this case, the discrete space of
 system excellence will be an analogical lattice.


 The  poset-like scales based on interval multiset estimates
 have been suggested in \cite{lev12a}.
%
 Analogically, two cases have to be considered:
 (i) system estimate by components,
 (ii) system estimate by components and by component
 compatibility.
%
%
%
 Fig. 7
 depicts
  the scale-poset and estimates for assessment problem
 \(P^{3,4}\) (assessment over scale \([1,3]\) with four elements;
 estimates  \((2,0,2)\), \((3,0,1)\), and \((1,0,3)\) are not used)
 \cite{lev12a}.
%
 Evidently, the above-mentioned resultant system ordinal scale \(D\)
  can used here as well.

\begin{center}
\begin{picture}(48,87)

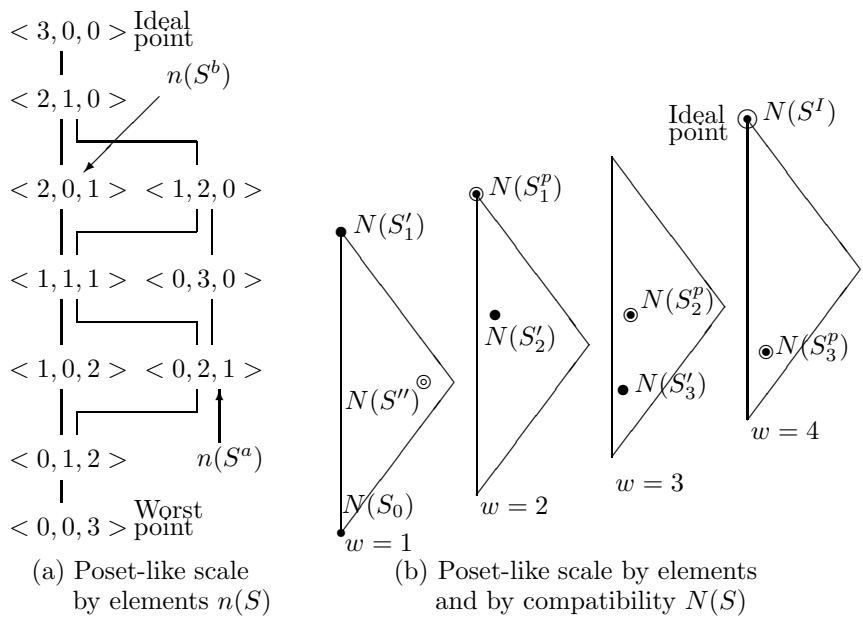
\put(05,00){\makebox(0,0)[bl] {Fig. 6. Poset-like scale for
 composite system based on ordinal estimates }}

\put(08,09){\makebox(0,0)[bl] {(a) Poset-like scale}}
\put(13.5,05){\makebox(0,0)[bl]{by elements \(n(S)\)}}

\put(05,81){\makebox(0,0)[bl]{\(<3,0,0>\) }}

\put(21.6,83){\makebox(0,0)[bl]{Ideal}}
\put(21.6,80){\makebox(0,0)[bl]{point}}

\put(12,77){\line(0,1){3}}
\put(05,72){\makebox(0,0)[bl]{\(<2,1,0>\)}}

\put(26,75){\makebox(0,0)[bl]{\(n(S^{b})\)}}
\put(25,74){\vector(-1,-1){10}}

\put(12,65){\line(0,1){6}}
\put(05,60){\makebox(0,0)[bl]{\(<2,0,1>\) }}

\put(12,53){\line(0,1){6}}
\put(05,48){\makebox(0,0)[bl]{\(<1,1,1>\) }}

\put(12,41){\line(0,1){6}}
\put(05,36){\makebox(0,0)[bl]{\(<1,0,2>\) }}


\put(12,29){\line(0,1){6}}
\put(05,24){\makebox(0,0)[bl]{\(<0,1,2>\) }}

\put(30,24){\makebox(0,0)[bl]{\(n(S^{a})\)}}
\put(33,28){\vector(0,1){7}}

\put(12,20){\line(0,1){3}}

\put(05,15){\makebox(0,0)[bl]{\(<0,0,3>\) }}

\put(21.6,18){\makebox(0,0)[bl]{Worst}}
\put(21.6,15){\makebox(0,0)[bl]{point}}

\put(14,68){\line(0,1){3}} \put(30,68){\line(-1,0){16}}
\put(30,65){\line(0,1){3}}

\put(23,60){\makebox(0,0)[bl]{\(<1,2,0>\) }}

\put(30,59){\line(0,-1){3}} \put(30,56){\line(-1,0){16}}
\put(14,56){\line(0,-1){3}}
\put(32,53){\line(0,1){6}}
\put(23,48){\makebox(0,0)[bl]{\(<0,3,0>\) }}

\put(14,44){\line(0,1){3}} \put(30,44){\line(-1,0){16}}
\put(30,41){\line(0,1){3}}

\put(32,41){\line(0,1){6}}
\put(23,36){\makebox(0,0)[bl]{\(<0,2,1>\) }}

\put(30,35){\line(0,-1){3}} \put(30,32){\line(-1,0){16}}
\put(14,32){\line(0,-1){3}}

\end{picture}
%
\begin{picture}(69,69)


\put(07,09){\makebox(0,0)[bl]{(b) Poset-like scale by elements}}
\put(12.5,05){\makebox(0,0)[bl]{and by compatibility \(N(S)\)}}

\put(00,16){\circle*{0.9}}
\put(0.1,17.7){\makebox(0,0)[bl]{\(N(S_{0})\)}}


\put(00,16){\line(0,1){40}} \put(00,16){\line(3,4){15}}
\put(00,56){\line(3,-4){15}}

\put(18,21){\line(0,1){40}} \put(18,21){\line(3,4){15}}
\put(18,61){\line(3,-4){15}}

\put(36,26){\line(0,1){40}} \put(36,26){\line(3,4){15}}
\put(36,66){\line(3,-4){15}}

\put(54,31){\line(0,1){40}} \put(54,31){\line(3,4){15}}
\put(54,71){\line(3,-4){15}}


\put(18,61){\circle*{1.1}} \put(18,61){\circle{1.9}}
\put(19.6,60){\makebox(0,0)[bl]{\(N(S^{p}_{1})\)}}

\put(38.5,45){\circle*{1.1}} \put(38.5,45){\circle{1.9}}
\put(39.8,45){\makebox(0,0)[bl]{\(N(S^{p}_{2})\)}}

\put(56.5,40){\circle*{1.1}} \put(56.5,40){\circle{1.9}}
\put(57.7,38.8){\makebox(0,0)[bl]{\(N(S^{p}_{3})\)}}


\put(00,56){\circle*{1.5}}
\put(01.6,55){\makebox(0,0)[bl]{\(N(S'_{1})\)}}

\put(20.5,45){\circle*{1.5}}
\put(19,40){\makebox(0,0)[bl]{\(N(S'_{2})\)}}

\put(37.5,35){\circle*{1.5}}
\put(39,33.8){\makebox(0,0)[bl]{\(N(S'_{3})\)}}


\put(11,36){\circle{0.75}} \put(11,36){\circle{1.7}}
\put(00.5,31.6){\makebox(0,0)[bl]{\(N(S'')\)}}





\put(54,71){\circle*{1}} \put(54,71){\circle{2.5}}

\put(43.5,70.5){\makebox(0,0)[bl]{Ideal}}
\put(43.5,67.5){\makebox(0,0)[bl]{point}}

\put(56,70){\makebox(0,0)[bl]{\(N(S^{I})\)}}

\put(00.5,13.5){\makebox(0,0)[bl]{\(w=1\)}}
\put(18.5,18.5){\makebox(0,0)[bl]{\(w=2\)}}
\put(36.5,21.5){\makebox(0,0)[bl]{\(w=3\)}}
\put(54.5,28.5){\makebox(0,0)[bl]{\(w=4\)}}

\end{picture}
\end{center}

\begin{center}
\begin{picture}(82,138)
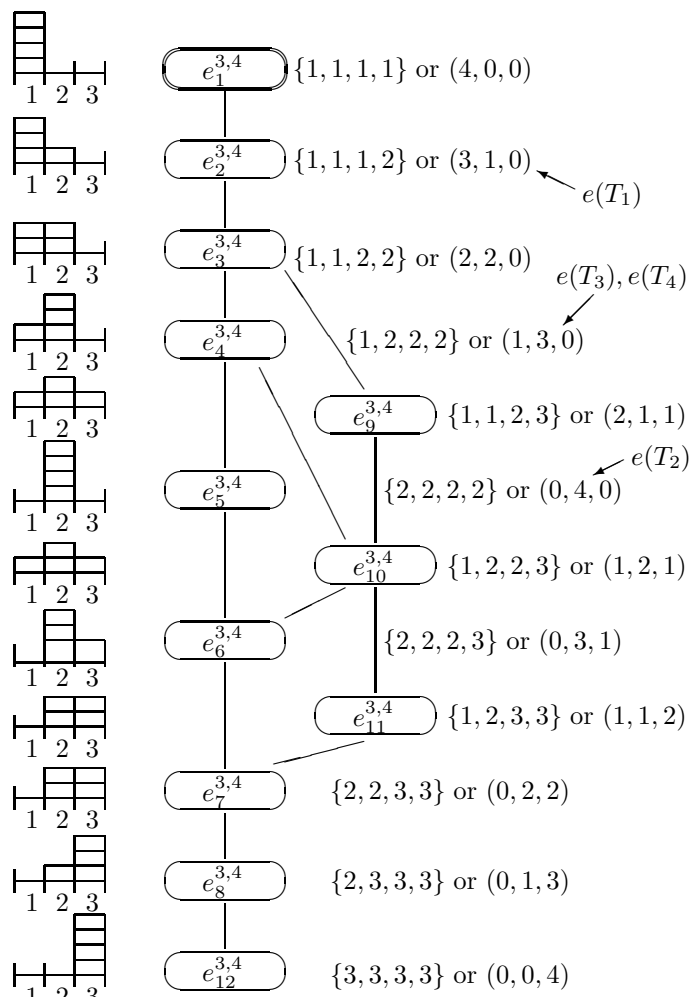
\put(016,00){\makebox(0,0)[bl] {Fig. 7. Poset-like scale
 (\(P^{3,4}\)) \cite{lev12a}
}}




\put(25,126.7){\makebox(0,0)[bl]{\(e^{3,4}_{1}\) }}

\put(28,129){\oval(16,5)} \put(28,129){\oval(16.5,5.5)}


\put(37,127){\makebox(0,0)[bl]{\(\{1,1,1,1\}\) or \((4,0,0)\) }}

\put(00,128.5){\line(0,1){08}} \put(04,128.5){\line(0,1){08}}

\put(00,130.5){\line(1,0){4}} \put(00,132.5){\line(1,0){4}}
\put(00,134.5){\line(1,0){4}} \put(00,136.5){\line(1,0){4}}

\put(00,128.5){\line(1,0){12}}

\put(00,127){\line(0,1){3}} \put(04,127){\line(0,1){3}}
\put(08,127){\line(0,1){3}} \put(12,127){\line(0,1){3}}

\put(01.5,124.5){\makebox(0,0)[bl]{\(1\)}}
\put(05.5,124.5){\makebox(0,0)[bl]{\(2\)}}
\put(09.5,124.5){\makebox(0,0)[bl]{\(3\)}}


\put(28,120){\line(0,1){6}}


\put(25,114.7){\makebox(0,0)[bl]{\(e^{3,4}_{2}\) }}

\put(28,117){\oval(16,5)}


\put(37,115){\makebox(0,0)[bl]{\(\{1,1,1,2\}\) or \((3,1,0)\) }}


\put(75.5,110.5){\makebox(0,0)[bl]{\(e(T_{1})\) }}
\put(74.5,113){\vector(-2,1){5}}


\put(00,116.5){\line(0,1){06}} \put(04,116.5){\line(0,1){06}}
\put(08,116.5){\line(0,1){02}}

\put(00,118.5){\line(1,0){8}} \put(00,120.5){\line(1,0){4}}
\put(00,122.5){\line(1,0){4}}

\put(00,116.5){\line(1,0){12}}

\put(00,115){\line(0,1){3}} \put(04,115){\line(0,1){3}}
\put(08,115){\line(0,1){3}} \put(12,115){\line(0,1){3}}

\put(01.5,112.5){\makebox(0,0)[bl]{\(1\)}}
\put(05.5,112.5){\makebox(0,0)[bl]{\(2\)}}
\put(09.5,112.5){\makebox(0,0)[bl]{\(3\)}}


\put(28,108){\line(0,1){6}}


\put(25,102.7){\makebox(0,0)[bl]{\(e^{3,4}_{3}\) }}

\put(28,105){\oval(16,5)}


\put(37,102){\makebox(0,0)[bl]{\(\{1,1,2,2\}\) or \((2,2,0)\) }}

\put(00,104.5){\line(0,1){04}}

\put(04,104.5){\line(0,1){04}} \put(08,104.5){\line(0,1){04}}
\put(00,106.5){\line(1,0){8}} \put(00,108.5){\line(1,0){8}}

\put(00,104.5){\line(1,0){12}}

\put(00,103){\line(0,1){3}} \put(04,103){\line(0,1){3}}
\put(08,103){\line(0,1){3}} \put(12,103){\line(0,1){3}}

\put(01.5,100.5){\makebox(0,0)[bl]{\(1\)}}
\put(05.5,100.5){\makebox(0,0)[bl]{\(2\)}}
\put(09.5,100.5){\makebox(0,0)[bl]{\(3\)}}


\put(28,96){\line(0,1){6}}


\put(25,90.7){\makebox(0,0)[bl]{\(e^{3,4}_{4}\) }}

\put(28,93){\oval(16,5)}



\put(72,99.5){\makebox(0,0)[bl]{\(e(T_{3}),e(T_{4})\) }}
\put(77,99){\vector(-1,-1){4}}


\put(44,91){\makebox(0,0)[bl]{\(\{1,2,2,2\}\) or \((1,3,0)\) }}

\put(00,93){\line(0,1){02}}

\put(04,93){\line(0,1){06}} \put(08,93){\line(0,1){06}}

\put(00,95){\line(1,0){8}} \put(04,97){\line(1,0){4}}
\put(04,99){\line(1,0){4}}

\put(00,93){\line(1,0){12}}

\put(00,91.5){\line(0,1){3}} \put(04,91.5){\line(0,1){3}}
\put(08,91.5){\line(0,1){3}} \put(12,91.5){\line(0,1){3}}

\put(01.5,89){\makebox(0,0)[bl]{\(1\)}}
\put(05.5,89){\makebox(0,0)[bl]{\(2\)}}
\put(09.5,89){\makebox(0,0)[bl]{\(3\)}}


\put(28,76){\line(0,1){14}}

\put(25,70.7){\makebox(0,0)[bl]{\(e^{3,4}_{5}\) }}

\put(28,73){\oval(16,5)}



\put(82,75.5){\makebox(0,0)[bl]{\(e(T_{2})\) }}
\put(81,77){\vector(-2,-1){4}}


\put(49,71){\makebox(0,0)[bl]{\(\{2,2,2,2\}\) or \((0,4,0)\) }}

\put(04,71.5){\line(0,1){08}} \put(08,71.5){\line(0,1){08}}

\put(04,73.5){\line(1,0){4}} \put(04,75.5){\line(1,0){4}}
\put(04,77.5){\line(1,0){4}} \put(04,79.5){\line(1,0){4}}

\put(00,71.5){\line(1,0){12}}

\put(00,70){\line(0,1){3}} \put(04,70){\line(0,1){3}}
\put(08,70){\line(0,1){3}} \put(12,70){\line(0,1){3}}

\put(01.5,67.5){\makebox(0,0)[bl]{\(1\)}}
\put(05.5,67.5){\makebox(0,0)[bl]{\(2\)}}
\put(09.5,67.5){\makebox(0,0)[bl]{\(3\)}}


\put(28,56){\line(0,1){14}}

\put(25,50.7){\makebox(0,0)[bl]{\(e^{3,4}_{6}\) }}

\put(28,53){\oval(16,5)}


\put(49,51){\makebox(0,0)[bl]{\(\{2,2,2,3\}\) or \((0,3,1)\) }}

\put(04,51){\line(0,1){06}} \put(08,51){\line(0,1){06}}
 \put(12,51){\line(0,1){02}}

\put(04,53){\line(1,0){8}} \put(04,55){\line(1,0){4}}
\put(04,57){\line(1,0){4}}

\put(00,50){\line(1,0){12}}

\put(00,49.5){\line(0,1){3}} \put(04,49.5){\line(0,1){3}}
\put(08,49.5){\line(0,1){3}} \put(12,49.5){\line(0,1){3}}

\put(01.5,47){\makebox(0,0)[bl]{\(1\)}}
\put(05.5,47){\makebox(0,0)[bl]{\(2\)}}
\put(09.5,47){\makebox(0,0)[bl]{\(3\)}}


\put(28,36){\line(0,1){14}}

\put(25,30.7){\makebox(0,0)[bl]{\(e^{3,4}_{7}\) }}

\put(28,33){\oval(16,5)}


\put(42,31){\makebox(0,0)[bl]{\(\{2,2,3,3\}\) or \((0,2,2)\) }}

\put(04,32){\line(0,1){04}}

\put(08,32){\line(0,1){04}} \put(12,32){\line(0,1){04}}
\put(04,34){\line(1,0){8}} \put(04,36){\line(1,0){8}}

\put(00,32){\line(1,0){12}}

\put(00,30.5){\line(0,1){3}} \put(04,30.5){\line(0,1){3}}
\put(08,30.5){\line(0,1){3}} \put(12,30.5){\line(0,1){3}}

\put(01.5,28){\makebox(0,0)[bl]{\(1\)}}
\put(05.5,28){\makebox(0,0)[bl]{\(2\)}}
\put(09.5,28){\makebox(0,0)[bl]{\(3\)}}


\put(28,24){\line(0,1){6}}

\put(25,18.7){\makebox(0,0)[bl]{\(e^{3,4}_{8}\) }}

\put(28,21){\oval(16,5)}


\put(42,19){\makebox(0,0)[bl]{\(\{2,3,3,3\}\) or \((0,1,3)\) }}

\put(04,21){\line(0,1){02}}

\put(08,21){\line(0,1){06}} \put(12,21){\line(0,1){06}}

\put(04,23){\line(1,0){8}} \put(08,25){\line(1,0){4}}
\put(08,27){\line(1,0){4}}

\put(00,21){\line(1,0){12}}

\put(00,19.5){\line(0,1){3}} \put(04,19.5){\line(0,1){3}}
\put(08,19.5){\line(0,1){3}} \put(12,19.5){\line(0,1){3}}

\put(01.5,17){\makebox(0,0)[bl]{\(1\)}}
\put(05.5,17){\makebox(0,0)[bl]{\(2\)}}
\put(09.5,17){\makebox(0,0)[bl]{\(3\)}}


\put(28,12){\line(0,1){6}}


\put(25,06.7){\makebox(0,0)[bl]{\(e^{3,4}_{12}\) }}

\put(28,09){\oval(16,5)}


\put(42,6.5){\makebox(0,0)[bl]{\(\{3,3,3,3\}\) or \((0,0,4)\) }}

\put(08,8.5){\line(0,1){08}} \put(12,8.5){\line(0,1){08}}
\put(08,10.5){\line(1,0){4}} \put(08,12.5){\line(1,0){4}}
\put(08,14.5){\line(1,0){4}} \put(08,16.5){\line(1,0){4}}

\put(00,8.5){\line(1,0){12}}

\put(00,07){\line(0,1){3}} \put(04,07){\line(0,1){3}}
\put(08,07){\line(0,1){3}} \put(12,07){\line(0,1){3}}

\put(01.5,4.5){\makebox(0,0)[bl]{\(1\)}}
\put(05.5,4.5){\makebox(0,0)[bl]{\(2\)}}
\put(09.5,4.5){\makebox(0,0)[bl]{\(3\)}}


\put(46.5,86.5){\line(-2,3){10.5}}




\put(45,80.7){\makebox(0,0)[bl]{\(e^{3,4}_{9}\) }}

\put(48,83){\oval(16,5)}


\put(57.5,81){\makebox(0,0)[bl]{\(\{1,1,2,3\}\) or \((2,1,1)\) }}

\put(00,84){\line(0,1){02}} \put(04,84){\line(0,1){04}}
\put(08,84){\line(0,1){04}} \put(12,84){\line(0,1){02}}

\put(00,86){\line(1,0){12}} \put(04,88){\line(1,0){4}}

\put(00,84){\line(1,0){12}}

\put(00,82.5){\line(0,1){3}} \put(04,82.5){\line(0,1){3}}
\put(08,82.5){\line(0,1){3}} \put(12,82.5){\line(0,1){3}}

\put(01.5,80){\makebox(0,0)[bl]{\(1\)}}
\put(05.5,80){\makebox(0,0)[bl]{\(2\)}}
\put(09.5,80){\makebox(0,0)[bl]{\(3\)}}


\put(44,66.5){\line(-1,2){11.4}}


\put(44,60){\line(-2,-1){8}}

\put(48,66){\line(0,1){14}}


\put(45,60.7){\makebox(0,0)[bl]{\(e^{3,4}_{10}\) }}

\put(48,63){\oval(16,5)}


\put(57.5,61){\makebox(0,0)[bl]{\(\{1,2,2,3\}\) or \((1,2,1)\) }}

\put(00,62){\line(0,1){02}} \put(04,62){\line(0,1){04}}
\put(08,62){\line(0,1){04}} \put(12,62){\line(0,1){02}}

\put(00,64){\line(1,0){12}} \put(04,66){\line(1,0){4}}

\put(00,62){\line(1,0){12}}

\put(00,60.5){\line(0,1){3}} \put(04,60.5){\line(0,1){3}}
\put(08,60.5){\line(0,1){3}} \put(12,60.5){\line(0,1){3}}

\put(01.5,58){\makebox(0,0)[bl]{\(1\)}}
\put(05.5,58){\makebox(0,0)[bl]{\(2\)}}
\put(09.5,58){\makebox(0,0)[bl]{\(3\)}}



\put(46.5,39.5){\line(-4,-1){12}}

\put(48,46){\line(0,1){14}}


\put(45,40.7){\makebox(0,0)[bl]{\(e^{3,4}_{11}\) }}

\put(48,43){\oval(16,5)}


\put(57.5,41){\makebox(0,0)[bl]{\(\{1,2,3,3\}\) or \((1,1,2)\) }}

\put(04,41.5){\line(0,1){04}}

\put(08,41.5){\line(0,1){04}} \put(12,41.5){\line(0,1){04}}
\put(04,43.5){\line(1,0){8}} \put(04,45.5){\line(1,0){8}}

\put(00,41.5){\line(1,0){12}}

\put(00,40){\line(0,1){3}} \put(04,40){\line(0,1){3}}
\put(08,40){\line(0,1){3}} \put(12,40){\line(0,1){3}}

\put(01.5,37.5){\makebox(0,0)[bl]{\(1\)}}
\put(05.5,37.5){\makebox(0,0)[bl]{\(2\)}}
\put(09.5,37.5){\makebox(0,0)[bl]{\(3\)}}


\end{picture}
\end{center}

 An example of four-component system composition
 is presented in Fig. 8.
%
 It is assumed,
 interval multiset estimates (scale from Fig. 7)
 are used for assessment of DAs.
%
%
%
%
 For evaluation of the final system consisting
 of four components,
 it is necessary to take into account estimates of compatibility
 (e.g., \([0,1,2,3]\)).
%
 The corresponding integrated poset-like scale is depicted in Fig. 9
 (median-like integral system estimates are assumed \cite{lev12a}).

\begin{center}
\begin{picture}(89,49)
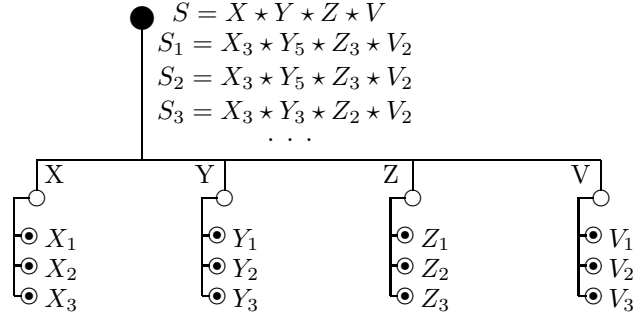
\put(07,00){\makebox(0,0)[bl] {Fig. 8. Example of
 four-component system}}

\put(4,05){\makebox(0,8)[bl]{\(X_{3}\)}}
\put(4,09){\makebox(0,8)[bl]{\(X_{2}\)}}
\put(4,13){\makebox(0,8)[bl]{\(X_{1}\)}}

\put(29,05){\makebox(0,8)[bl]{\(Y_{3}\)}}
\put(29,09){\makebox(0,8)[bl]{\(Y_{2}\)}}
\put(29,13){\makebox(0,8)[bl]{\(Y_{1}\)}}

\put(54,05){\makebox(0,8)[bl]{\(Z_{3}\)}}
\put(54,09){\makebox(0,8)[bl]{\(Z_{2}\)}}
\put(54,13){\makebox(0,8)[bl]{\(Z_{1}\)}}

\put(79,05){\makebox(0,8)[bl]{\(V_{3}\)}}
\put(79,09){\makebox(0,8)[bl]{\(V_{2}\)}}
\put(79,13){\makebox(0,8)[bl]{\(V_{1}\)}}






\put(3,20){\circle{2}} \put(28,20){\circle{2}}
\put(53,20){\circle{2}} \put(78,20){\circle{2}}

\put(0,20){\line(1,0){02}} \put(25,20){\line(1,0){02}}
\put(50,20){\line(1,0){02}} \put(75,20){\line(1,0){02}}

\put(0,20){\line(0,-1){13}} \put(25,20){\line(0,-1){13}}
\put(50,20){\line(0,-1){13}} \put(75,20){\line(0,-1){13}}

\put(75,07){\line(1,0){01}} \put(75,11){\line(1,0){01}}
\put(75,15){\line(1,0){01}}


\put(77,15){\circle{2}} \put(77,15){\circle*{1}}
\put(77,11){\circle{2}} \put(77,11){\circle*{1}}
\put(77,07){\circle{2}} \put(77,07){\circle*{1}}

\put(50,15){\line(1,0){01}} \put(50,11){\line(1,0){01}}
\put(50,07){\line(1,0){01}}

\put(52,15){\circle{2}} \put(52,15){\circle*{1}}
\put(52,11){\circle{2}} \put(52,11){\circle*{1}}
\put(52,07){\circle{2}} \put(52,07){\circle*{1}}

\put(25,07){\line(1,0){01}} \put(25,11){\line(1,0){01}}
\put(25,15){\line(1,0){01}}


\put(27,15){\circle{2}} \put(27,15){\circle*{1}}
\put(27,11){\circle{2}} \put(27,11){\circle*{1}}
\put(27,07){\circle{2}} \put(27,07){\circle*{1}}

\put(0,07){\line(1,0){01}} \put(0,11){\line(1,0){01}}
\put(0,15){\line(1,0){01}}

\put(2,11){\circle{2}} \put(2,15){\circle{2}}
\put(2,11){\circle*{1}} \put(2,15){\circle*{1}}
\put(2,07){\circle{2}} \put(2,07){\circle*{1}}
\put(3,25){\line(0,-1){04}} \put(28,25){\line(0,-1){04}}
\put(53,25){\line(0,-1){04}} \put(78,25){\line(0,-1){04}}

\put(3,25){\line(1,0){75}} \put(17,25){\line(0,1){19}}

\put(17,44){\circle*{3}}

\put(4,22){\makebox(0,8)[bl]{X}} \put(24,22){\makebox(0,8)[bl]{Y}}
\put(49,22){\makebox(0,8)[bl]{Z}}
\put(74,22){\makebox(0,8)[bl]{V}}

\put(21,43.5){\makebox(0,8)[bl]{\(S = X \star Y \star Z \star
V\)}}

\put(19,39){\makebox(0,8)[bl] {\(S_{1}=X_{3}\star Y_{5}\star Z_{3}
 \star V_{2}\)}}

\put(19,34.5){\makebox(0,8)[bl] {\(S_{2}=X_{3}\star Y_{5}\star
Z_{3}
 \star V_{2}\)}}

\put(19,30){\makebox(0,8)[bl] {\(S_{3}=X_{3}\star Y_{3}\star Z_{2}
\star V_{2}\)}}

\put(32.5,27.6){\makebox(0,8)[bl] { . . .}}





\end{picture}
\end{center}

\begin{center}
\begin{picture}(41,140)
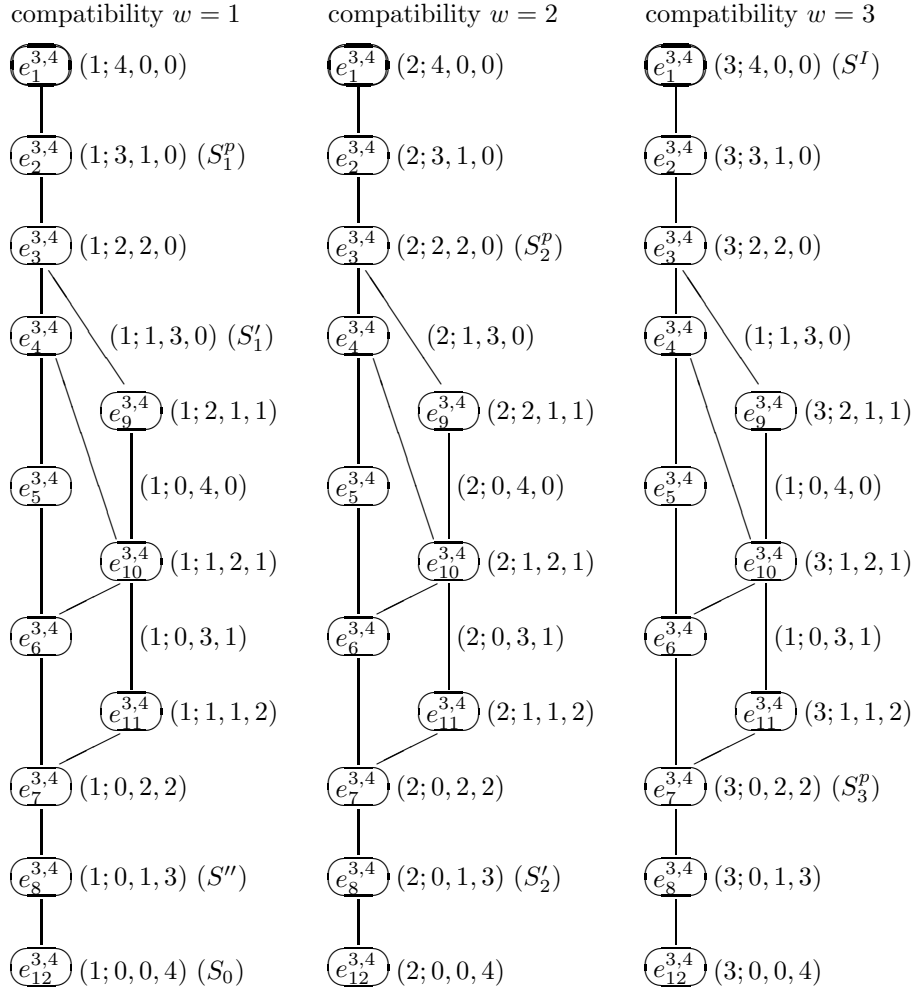
\put(00,00){\makebox(0,0)[bl] {Fig. 9. Integrated poset
 (assessment problem \(P^{3,4}\)),
  compatibility scale \([1,2,3]\)}}

\put(00,134){\makebox(0,0)[bl]{compatibility \(w=1\) }}



\put(01,126.7){\makebox(0,0)[bl]{\(e^{3,4}_{1}\) }}

\put(04,129){\oval(08,5)} \put(04,129){\oval(7.5,5.5)}


\put(09,127){\makebox(0,0)[bl]{\((1;4,0,0)\) }}


\put(04,120){\line(0,1){6}}


\put(01,114.7){\makebox(0,0)[bl]{\(e^{3,4}_{2}\) }}

\put(04,117){\oval(08,5)}


\put(09,115){\makebox(0,0)[bl]{\((1;3,1,0) ~(S^{p}_{1})\) }}


\put(04,108){\line(0,1){6}}


\put(01,102.7){\makebox(0,0)[bl]{\(e^{3,4}_{3}\) }}

\put(04,105){\oval(08,5)}


\put(09,103){\makebox(0,0)[bl]{\((1;2,2,0)\) }}


\put(04,96){\line(0,1){6}}


\put(01,90.7){\makebox(0,0)[bl]{\(e^{3,4}_{4}\) }}

\put(04,93){\oval(08,5)}


\put(13,91){\makebox(0,0)[bl]{\((1;1,3,0) ~(S'_{1})\) }}


\put(04,76){\line(0,1){14}}

\put(01,70.7){\makebox(0,0)[bl]{\(e^{3,4}_{5}\) }}

\put(04,73){\oval(08,5)}


\put(17,71){\makebox(0,0)[bl]{\((1;0,4,0)\) }}


\put(04,56){\line(0,1){14}}

\put(01,50.7){\makebox(0,0)[bl]{\(e^{3,4}_{6}\) }}

\put(04,53){\oval(08,5)}


\put(17,51){\makebox(0,0)[bl]{\((1;0,3,1)\) }}


\put(04,36){\line(0,1){14}}

\put(01,30.7){\makebox(0,0)[bl]{\(e^{3,4}_{7}\) }}

\put(04,33){\oval(08,5)}


\put(09,31){\makebox(0,0)[bl]{\((1;0,2,2)\) }}


\put(04,24){\line(0,1){6}}

\put(01,18.7){\makebox(0,0)[bl]{\(e^{3,4}_{8}\) }}

\put(04,21){\oval(08,5)}


\put(09,19){\makebox(0,0)[bl]{\((1;0,1,3) ~(S'')\) }}



\put(04,12){\line(0,1){6}}


\put(01,06.7){\makebox(0,0)[bl]{\(e^{3,4}_{12}\) }}

\put(04,09){\oval(08,5)}


\put(09,6.5){\makebox(0,0)[bl]{\((1;0,0,4) ~(S_{0})\) }}



\put(15,86.6){\line(-2,3){10}}

\put(13,80.7){\makebox(0,0)[bl]{\(e^{3,4}_{9}\) }}

\put(16,83){\oval(08,5)}


\put(21,81){\makebox(0,0)[bl]{\((1;2,1,1)\) }}


\put(14,66){\line(-1,3){8}}

\put(14.5,60){\line(-2,-1){8}}

\put(16,66){\line(0,1){14}}

\put(13,60.7){\makebox(0,0)[bl]{\(e^{3,4}_{10}\) }}

\put(16,63){\oval(08,5)}


\put(21,61){\makebox(0,0)[bl]{\((1;1,2,1)\) }}


\put(14.5,40){\line(-2,-1){8}}

\put(16,46){\line(0,1){14}}

\put(13,40.7){\makebox(0,0)[bl]{\(e^{3,4}_{11}\) }}

\put(16,43){\oval(08,5)}


\put(21,41){\makebox(0,0)[bl]{\((1;1,1,2)\) }}


\end{picture}
%
\begin{picture}(41,140)

\put(00,134){\makebox(0,0)[bl]{compatibility \(w=2\) }}




\put(01,126.7){\makebox(0,0)[bl]{\(e^{3,4}_{1}\) }}

\put(04,129){\oval(08,5)} \put(04,129){\oval(7.5,5.5)}


\put(09,127){\makebox(0,0)[bl]{\((2;4,0,0)\) }}


\put(04,120){\line(0,1){6}}


\put(01,114.7){\makebox(0,0)[bl]{\(e^{3,4}_{2}\) }}

\put(04,117){\oval(08,5)}


\put(09,115){\makebox(0,0)[bl]{\((2;3,1,0)\) }}


\put(04,108){\line(0,1){6}}


\put(01,102.7){\makebox(0,0)[bl]{\(e^{3,4}_{3}\) }}

\put(04,105){\oval(08,5)}


\put(09,103){\makebox(0,0)[bl]{\((2;2,2,0) ~(S^{p}_{2})\) }}


\put(04,96){\line(0,1){6}}


\put(01,90.7){\makebox(0,0)[bl]{\(e^{3,4}_{4}\) }}

\put(04,93){\oval(08,5)}


\put(13,91){\makebox(0,0)[bl]{\((2;1,3,0)\) }}


\put(04,76){\line(0,1){14}}

\put(01,70.7){\makebox(0,0)[bl]{\(e^{3,4}_{5}\) }}

\put(04,73){\oval(08,5)}


\put(17,71){\makebox(0,0)[bl]{\((2;0,4,0)\) }}


\put(04,56){\line(0,1){14}}

\put(01,50.7){\makebox(0,0)[bl]{\(e^{3,4}_{6}\) }}

\put(04,53){\oval(08,5)}


\put(17,51){\makebox(0,0)[bl]{\((2;0,3,1)\) }}


\put(04,36){\line(0,1){14}}

\put(01,30.7){\makebox(0,0)[bl]{\(e^{3,4}_{7}\) }}

\put(04,33){\oval(08,5)}


\put(09,31){\makebox(0,0)[bl]{\((2;0,2,2)\) }}


\put(04,24){\line(0,1){6}}

\put(01,18.7){\makebox(0,0)[bl]{\(e^{3,4}_{8}\) }}

\put(04,21){\oval(08,5)}


\put(09,19){\makebox(0,0)[bl]{\((2;0,1,3) ~(S'_{2})\) }}


\put(04,12){\line(0,1){6}}


\put(01,06.7){\makebox(0,0)[bl]{\(e^{3,4}_{12}\) }}

\put(04,09){\oval(08,5)}


\put(09,6.5){\makebox(0,0)[bl]{\((2;0,0,4)\) }}


\put(15,86.6){\line(-2,3){10}}

\put(13,80.7){\makebox(0,0)[bl]{\(e^{3,4}_{9}\) }}

\put(16,83){\oval(08,5)}


\put(21,81){\makebox(0,0)[bl]{\((2;2,1,1)\) }}


\put(14,66){\line(-1,3){8}}

\put(14.5,60){\line(-2,-1){8}}

\put(16,66){\line(0,1){14}}

\put(13,60.7){\makebox(0,0)[bl]{\(e^{3,4}_{10}\) }}

\put(16,63){\oval(08,5)}


\put(21,61){\makebox(0,0)[bl]{\((2;1,2,1)\) }}


\put(14.5,40){\line(-2,-1){8}}

\put(16,46){\line(0,1){14}}

\put(13,40.7){\makebox(0,0)[bl]{\(e^{3,4}_{11}\) }}

\put(16,43){\oval(08,5)}


\put(21,41){\makebox(0,0)[bl]{\((2;1,1,2)\) }}


\end{picture}
%
\begin{picture}(35,140)

\put(00,134){\makebox(0,0)[bl]{compatibility \(w=3\) }}




\put(01,126.7){\makebox(0,0)[bl]{\(e^{3,4}_{1}\) }}

\put(04,129){\oval(08,5)} \put(04,129){\oval(7.5,5.5)}


\put(09,127){\makebox(0,0)[bl]{\((3;4,0,0) ~(S^{I})\) }}


\put(04,120){\line(0,1){6}}


\put(01,114.7){\makebox(0,0)[bl]{\(e^{3,4}_{2}\) }}

\put(04,117){\oval(08,5)}


\put(09,115){\makebox(0,0)[bl]{\((3;3,1,0)\) }}


\put(04,108){\line(0,1){6}}


\put(01,102.7){\makebox(0,0)[bl]{\(e^{3,4}_{3}\) }}

\put(04,105){\oval(08,5)}


\put(09,103){\makebox(0,0)[bl]{\((3;2,2,0)\) }}


\put(04,96){\line(0,1){6}}


\put(01,90.7){\makebox(0,0)[bl]{\(e^{3,4}_{4}\) }}

\put(04,93){\oval(08,5)}


\put(13,91){\makebox(0,0)[bl]{\((1;1,3,0)\) }}


\put(04,76){\line(0,1){14}}

\put(01,70.7){\makebox(0,0)[bl]{\(e^{3,4}_{5}\) }}

\put(04,73){\oval(08,5)}


\put(17,71){\makebox(0,0)[bl]{\((1;0,4,0)\) }}


\put(04,56){\line(0,1){14}}

\put(01,50.7){\makebox(0,0)[bl]{\(e^{3,4}_{6}\) }}

\put(04,53){\oval(08,5)}


\put(17,51){\makebox(0,0)[bl]{\((1;0,3,1)\) }}


\put(04,36){\line(0,1){14}}

\put(01,30.7){\makebox(0,0)[bl]{\(e^{3,4}_{7}\) }}

\put(04,33){\oval(08,5)}


\put(09,31){\makebox(0,0)[bl]{\((3;0,2,2) ~(S^{p}_{3})\) }}


\put(04,24){\line(0,1){6}}

\put(01,18.7){\makebox(0,0)[bl]{\(e^{3,4}_{8}\) }}

\put(04,21){\oval(08,5)}


\put(09,19){\makebox(0,0)[bl]{\((3;0,1,3)\) }}


\put(04,12){\line(0,1){6}}


\put(01,06.7){\makebox(0,0)[bl]{\(e^{3,4}_{12}\) }}

\put(04,09){\oval(08,5)}


\put(09,6.5){\makebox(0,0)[bl]{\((3;0,0,4)\) }}


\put(15,86.6){\line(-2,3){10}}

\put(13,80.7){\makebox(0,0)[bl]{\(e^{3,4}_{9}\) }}

\put(16,83){\oval(08,5)}


\put(21,81){\makebox(0,0)[bl]{\((3;2,1,1)\) }}


\put(14,66){\line(-1,3){8}}

\put(14.5,60){\line(-2,-1){8}}

\put(16,66){\line(0,1){14}}

\put(13,60.7){\makebox(0,0)[bl]{\(e^{3,4}_{10}\) }}

\put(16,63){\oval(08,5)}


\put(21,61){\makebox(0,0)[bl]{\((3;1,2,1)\) }}


\put(14.5,40){\line(-2,-1){8}}

\put(16,46){\line(0,1){14}}

\put(13,40.7){\makebox(0,0)[bl]{\(e^{3,4}_{11}\) }}

\put(16,43){\oval(08,5)}


\put(21,41){\makebox(0,0)[bl]{\((3;1,1,2)\) }}


\end{picture}
\end{center}

 Further, an illustration of the resultant system ordinal scale \(D\)
 is the following (Fig. 9):


  {\it 1.} The ideal solution  \(S^{I}\),
  estimate is: \(e(S^{I}) = (3;4,0,0)\).

  {\it 2.} A layer of Pareto-efficient solutions:
  \( \{ S_{1}^{p}, S_{2}^{p}, S_{3}^{p} \} \);
 estimates (points) are:
 \(e(S^{p}_{1}) = (1;3,1,0)\),
 \(e(S^{p}_{2})=(2;2,2,0)\), and
 \(e(S^{p}_{3})=(3;0,2,2)\).

  {\it 3.} A next layer of quality
 (e.g., neighborhood of Pareto-efficient solutions  layer):
 \( \{ S'_{1}, S'_{2} \} \);
 estimates (points) are:
  \(e(S'_{1}) = (1;1,3,0)\), and
 \(e(S'_{2})=(2;0,1,3)\).


  {\it 4.} A next layer of quality \(S''\);
    estimate is:  \(e(S'')=(1;0,1,3)\).

 {\it 5.} The worst solution \(S_{0}\);
 estimate is: \(e(S_{0}) = (1;0,0,3)\).

\section{Transformation of Scales}

 Generally, main transformation problems for basic assessment  scales
 are shown in Table 2
  (note, resultant ordinal scale corresponds often to final solutions).

\begin{center}
\begin{picture}(117,117)

\put(14,113){\makebox(0,0)[bl]{Table 2. Examined problems for
 transformation of scales}}

\put(00,0){\line(1,0){117}} \put(00,87){\line(1,0){117}}
\put(24,105){\line(1,0){93}} \put(00,111){\line(1,0){117}}


\put(00,0){\line(0,1){111}} \put(24,0){\line(0,1){111}}

\put(38,0){\line(0,1){105}} \put(57,0){\line(0,1){105}}
\put(76,0){\line(0,1){105}} \put(98,0){\line(0,1){105}}

\put(117,0){\line(0,1){111}}


\put(01,107){\makebox(0,0)[bl]{Initial}}
\put(01,104){\makebox(0,0)[bl]{scale}}


\put(57,107){\makebox(0,0)[bl]{Resultant scale}}


\put(25,100.5){\makebox(0,0)[bl]{Quanti-}}
\put(25,98){\makebox(0,0)[bl]{tative}}
\put(25,95){\makebox(0,0)[bl]{scale}}

\put(39,101){\makebox(0,0)[bl]{Ordinal }}
\put(39,98){\makebox(0,0)[bl]{scale}}

\put(58,101){\makebox(0,0)[bl]{Multi-}}
\put(58,98){\makebox(0,0)[bl]{criteria}}
\put(58,94){\makebox(0,0)[bl]{description}}

\put(77,101){\makebox(0,0)[bl]{Poset-like }}
\put(77,98){\makebox(0,0)[bl]{scale based  }}
\put(77,95){\makebox(0,0)[bl]{on ordinal}}
\put(77,92){\makebox(0,0)[bl]{estimates}}

\put(99,101){\makebox(0,0)[bl]{Poset-like }}
\put(99,98){\makebox(0,0)[bl]{scale based}}
\put(99,95){\makebox(0,0)[bl]{on interval}}
\put(99,92){\makebox(0,0)[bl]{multiset}}
\put(99,89){\makebox(0,0)[bl]{estimates}}


\put(01,82){\makebox(0,0)[bl]{1.Quantitative}}
\put(03,78.5){\makebox(0,0)[bl]{scale}}

\put(30,82){\makebox(0,0)[bl]{\(\star\)}}
\put(46,82){\makebox(0,0)[bl]{\(\star\)}}
\put(65,82){\makebox(0,0)[bl]{\(-\)}}
\put(85.5,82){\makebox(0,0)[bl]{\(-\)}}
\put(106,82){\makebox(0,0)[bl]{\(-\)}}


\put(01,73){\makebox(0,0)[bl]{2.Ordinal}}
\put(03,69.5){\makebox(0,0)[bl]{scale}}

\put(30,73){\makebox(0,0)[bl]{\(-\)}}
\put(46,73){\makebox(0,0)[bl]{\(\star\)}}
\put(65,73){\makebox(0,0)[bl]{\(-\)}}
\put(85.5,73){\makebox(0,0)[bl]{\(-\)}}
\put(106,73){\makebox(0,0)[bl]{\(-\)}}


\put(01,64){\makebox(0,0)[bl]{3.Multicriteria}}
\put(03,59.5){\makebox(0,0)[bl]{description }}
\put(03,55.2){\makebox(0,0)[bl]{(based on}}
\put(03,52){\makebox(0,0)[bl]{ordinal}}
\put(03,47.5){\makebox(0,0)[bl]{and/or}}
\put(03,44){\makebox(0,0)[bl]{quantitative}}
\put(03,40){\makebox(0,0)[bl]{estimates)}}

\put(30,64){\makebox(0,0)[bl]{\(\star\)}}
\put(25,60){\makebox(0,0)[bl]{\cite{fis70,kee76,ste86}}}

\put(46,64){\makebox(0,0)[bl]{\(\star\)}}
\put(40.5,60){\makebox(0,0)[bl]{\cite{lev98,levmih88,mirkin79}}}
\put(42,56){\makebox(0,0)[bl]{\cite{roy96,zap02}}}

\put(65,64){\makebox(0,0)[bl]{\(\star\)}}
\put(64,60){\makebox(0,0)[bl]{\cite{borg05}}}

\put(85.5,64){\makebox(0,0)[bl]{\(\star\)}}
\put(82,60){\makebox(0,0)[bl]{\cite{lev98,lev06}}}
\put(82,56){\makebox(0,0)[bl]{\cite{lev09,lev12morph}}}

\put(106,64){\makebox(0,0)[bl]{\(\star\)}}
\put(104,60){\makebox(0,0)[bl]{\cite{lev12a}}}


\put(01,35){\makebox(0,0)[bl]{4.Poset-like}}
\put(03,31){\makebox(0,0)[bl]{scale based}}
\put(03,27){\makebox(0,0)[bl]{on ordinal}}
\put(03,23){\makebox(0,0)[bl]{estimates}}

\put(30,35){\makebox(0,0)[bl]{\(-\)}}
\put(46,35){\makebox(0,0)[bl]{\(\star\)}}
\put(42.5,31){\makebox(0,0)[bl]{\cite{lev98,lev06}}}

\put(65,35){\makebox(0,0)[bl]{\(-\)}}
\put(85.5,35){\makebox(0,0)[bl]{\(-\)}}
\put(106,35){\makebox(0,0)[bl]{\(-\)}}


\put(01,18){\makebox(0,0)[bl]{5.Poset-like}}
\put(03,14){\makebox(0,0)[bl]{scale based}}
\put(03,10){\makebox(0,0)[bl]{on interval}}
\put(03,06){\makebox(0,0)[bl]{multiset }}
\put(03,02){\makebox(0,0)[bl]{estimates}}

\put(30,18){\makebox(0,0)[bl]{\(-\)}}
\put(46,18){\makebox(0,0)[bl]{\(\star\)}}
\put(44,14){\makebox(0,0)[bl]{\cite{lev12a}}}

\put(65,18){\makebox(0,0)[bl]{\(-\)}}
\put(85.5,18){\makebox(0,0)[bl]{\(-\)}}
\put(106,18){\makebox(0,0)[bl]{\(\star\)}}
\put(104,14){\makebox(0,0)[bl]{\cite{lev12a}}}

\end{picture}
\end{center}
 Here, the following basic scale transformation problems are
 considered:

 {\bf 1.} {\it Quantitative scale} \(\Rightarrow\)
 {\it Quantitative scale}.

 {\bf 2.} {\it Quantitative scale} \(\Rightarrow\)
 {\it Ordinal scale}.

 {\bf 3.} {\it Ordinal scale} \(\Rightarrow\)
 {\it Ordinal scale}.

 {\bf 4.} {\it Multicriteria description} \(\Rightarrow\)
 {\it Ordinal scale}.
 This is multicriteria ranking or sorting problem
 (e.g., \cite{lev98,levmih88,mirkin79,roy96,zap02}).

 {\bf 5.}  {\it Poset-like scale}  \(\Rightarrow\)
 {\it Ordinal scale}
 (e.g., \cite{lev98,lev06,lev12a}).
%

  {\bf 6.} {\it Multicriteria description} \(\Rightarrow\)
  {\it Quantitative scale}.
  This is decision making based on utility function analysis
 (e.g., \cite{fis70,kee76,ste86}).

   {\bf 7.} {\it Multicriteria description (ordinal scales)} \(\Rightarrow\)
  {\it Poset-like scale, based on ordinal estimates}.
 Here, the same ordinal scales are assumed (i.e., for each system part/component).
  This scale transformation type is described in
 (e.g., \cite{lev98,lev06,lev09,lev12morph}).

   {\bf 8.} {\it Multicriteria description (ordinal scales)} \(\Rightarrow\)
  {\it Poset-like scale, based on interval multiset estimates}.
 Here, the same ordinal scales are assumed (i.e., for each system part/component).
  This scale transformation type is described in
 (e.g., \cite{lev12a}).

   {\bf 9.} {\it Poset-like scale, based on interval multiset estimates} \(\Rightarrow\)
  {\it Poset-like scale, based on interval multiset estimates}.
  This scale transformation type is briefly described in
 (e.g., \cite{lev12a}).

  {\bf 10.}
 {\it Multicriteria description} \(\Rightarrow\)
 {\it Multicriteria description}.
 (some simple mappings, multidimensional scaling, etc.,
 e.g., \cite{borg05}).

 Note, the above-mentioned types 4, 6, 7, and 8
 correspond to the scale integration problem.
%


 The first type of transformation
 (i.e., {\it quantitative scale} \(\Rightarrow\) {\it quantitative scale})
 (Fig. 10)
 can be based on a linear function
 ~(\( y = a x + b \)).

\begin{center}
\begin{picture}(70,51)

\put(00,00){\makebox(0,0)[bl]{Fig. 10.
 Quantitative scale \(\Rightarrow\)  quantitative scale}}

\put(00,9){\makebox(0,0)[bl]{(a) Quantita-}}
\put(05.7,6){\makebox(0,0)[bl]{tive scale}}

\put(08.5,18){\line(1,0){3}}

\put(08,13.5){\makebox(0,0)[bl]{(\(0\))}}

\put(10,18){\vector(0,1){27.5}}



\put(10,39){\circle*{1.4}} \put(10,39){\circle{2.4}}

\put(00,41){\makebox(0,0)[bl]{Best}}
\put(00,38){\makebox(0,0)[bl]{point}}
\put(03,35.8){\makebox(0,0)[bl]{\(\alpha\)}}

\put(12,39){\line(1,0){21}} \put(33,39){\line(2,1){12}}
\put(45,45){\vector(1,0){11}}


\put(12,35){\vector(1,0){44}}


\put(12,27){\vector(1,0){44}}



\put(10,23){\circle*{1.5}}

\put(12,23){\line(1,0){21}} \put(33,23){\line(2,-1){6}}
\put(39,20){\vector(1,0){17}}

\put(00,26){\makebox(0,0)[bl]{Worst}}
\put(00,23){\makebox(0,0)[bl]{point}}
\put(03,20){\makebox(0,0)[bl]{\(\beta\)}}




\put(32,31){\makebox(0,0)[bl]{. . .}}


\put(46,9){\makebox(0,0)[bl]{(b) Quantita-}}
\put(51.7,6){\makebox(0,0)[bl]{tive scale}}

\put(57.5,18){\line(1,0){3}}

\put(57,13.5){\makebox(0,0)[bl]{(\(0\))}}

\put(59,18){\vector(0,1){31.5}}



\put(59,45){\circle*{1.4}} \put(59,45){\circle{2.4}}

\put(61,47){\makebox(0,0)[bl]{Best}}
\put(61,44){\makebox(0,0)[bl]{point}}
\put(63,41.8){\makebox(0,0)[bl]{\(\alpha\)}}


\put(59,20){\circle*{1.5}}

\put(61,23){\makebox(0,0)[bl]{Worst}}
\put(61,20){\makebox(0,0)[bl]{point}}
\put(63,17){\makebox(0,0)[bl]{\(\beta\)}}

\end{picture}
\end{center}

 The second  type of transformation
 (i.e., {\it quantitative scale} \(\Rightarrow\)
 {\it ordinal scale})
 is illustrated in Fig. 11.
 Here, the quantitative scale
 (or considered value interval \((\beta,\alpha)\))
 is divided into a set of interval,
 and each interval corresponds to a level of the resultant
 ordinal scale.
 The dividing procedure
 (i.e., definition of the thresholds)
 may be based on various approaches
 (e.g., computing scheme, expert judgment,
 usage of reference points)
 (e.g., \cite{alek76,larmf86,levmih88,ser83})

\begin{center}
\begin{picture}(76,55)

\put(02.5,00){\makebox(0,0)[bl]{Fig. 11.
 Quantitative scale \(\Rightarrow\) ordinal scale}}


\put(06,8){\makebox(0,0)[bl]{(a) Quantita-}}
\put(11.7,5){\makebox(0,0)[bl]{tive scale}}

\put(17.5,17){\line(1,0){3}}

\put(17,12.5){\makebox(0,0)[bl]{(\(0\))}}

\put(19,17){\vector(0,1){36.5}}


\put(00,37){\makebox(0,0)[bl]{Thre-}}
\put(00,34){\makebox(0,0)[bl]{sholds}}

\put(10,38){\line(1,1){6}}

\put(10,37){\line(4,1){6}}

\put(10,36){\line(4,-1){6}}

\put(10,35){\line(2,-3){6}}



\put(19,49){\circle*{1.4}} \put(19,49){\circle{2.4}}

\put(09,51){\makebox(0,0)[bl]{Best}}
\put(09,48){\makebox(0,0)[bl]{point}}
\put(12,45.8){\makebox(0,0)[bl]{\(\alpha\)}}

\put(20,46){\vector(1,0){37}}

\put(17.5,44){\line(1,0){3}} \put(17.5,44.12){\line(1,0){3}}

\put(20,41){\vector(1,0){37}}

\put(17.5,39){\line(1,0){3}} \put(17.5,39.1){\line(1,0){3}}

\put(20,36){\vector(1,0){37}}

\put(17.5,34){\line(1,0){3}} \put(17.5,34.1){\line(1,0){3}}


\put(20,29){\makebox(0,0)[bl]{...}}


\put(17.5,24.8){\line(1,0){3}} \put(17.5,24.9){\line(1,0){3}}


\put(20,22){\vector(1,0){37}}

\put(19,20){\circle*{1.5}}

\put(09,21.1){\makebox(0,0)[bl]{Worst}}
\put(09,18.1){\makebox(0,0)[bl]{point}}
\put(12,14.8){\makebox(0,0)[bl]{\(\beta\)}}


\put(36.5,29){\makebox(0,0)[bl]{. . .}}



\put(48,8){\makebox(0,0)[bl]{(b) Ordinal}}
\put(55,5){\makebox(0,0)[bl]{scale}}

\put(65,48){\makebox(0,0)[bl]{Best}}
\put(65,45){\makebox(0,0)[bl]{point}}

\put(60,46){\circle*{1.4}} \put(60,46){\circle{2.4}}
\put(62,45.5){\makebox(0,0)[bl]{\(1\)}}

\put(60,46){\vector(0,-1){4}}

\put(62,40.5){\makebox(0,0)[bl]{\(2\)}}

\put(60,41){\circle*{1.9}} \put(60,41){\vector(0,-1){4}}

\put(62,35.5){\makebox(0,0)[bl]{\(3\)}}

\put(60,36){\circle*{1.9}} \put(60,36){\vector(0,-1){4}}

\put(58.5,29){\makebox(0,0)[bl]{...}}


\put(60,27){\vector(0,-1){4}}


\put(62,21.5){\makebox(0,0)[bl]{\(\kappa\)}}

\put(60,22){\circle*{1.9}}

\put(65,22.5){\makebox(0,0)[bl]{Worst}}
\put(65,19.5){\makebox(0,0)[bl]{point}}

\end{picture}
\end{center}

 The third type  is the following.
 Two typical cases for transformation (i.e., mapping)
 {\it ordinal scale} \( \Rightarrow\) {\it ordinal scale}
 are depicted in Fig. 12.
 The mapping can be based on expert judgment
 (i.e., professional knowledge of domain experts).

 For the fourth type of the above-mentioned transformation
 (i.e., {\it multicriteria description} \(\Rightarrow\)
 {\it ordinal scale}),
 the following main approaches are used:

 (1) two-stage method:
 vector-estimates \(\Rightarrow\) utility function \(\Rightarrow\) resultant ordinal estimate,

 (2) series detection of Pareto-layers,

 (3) series detection of maximal points,


 (4) usage of dividing curves of equal quality,
 i.e., curves of equal quality or subdomains of equal quality;
 here, expert judgment procedures or logical methods can be used
 (e.g., \cite{levmih88})
 (Fig. 13),

 (5) frameworks based on
 analysis and usage of reference solutions,

 (6) outranking techniques (ELECTRE, PROMETHEE, etc.)
 (e.g., \cite{brans84,roy96}),

 (7) special interactive procedures
 based on logical methods
 (e.g., \cite{alek76,levmih88,ser83}),

 (8) usage of an ordinal scale \(D\)
 (e.g., Fig. 5c, Fig. 6b):~
 (i) the ideal solution,
 (ii) Pareto-efficient points,
 (iii) near Pareto-efficient points
 (the points are close to the Pareto-layer),
 (iv) some other points,
 (v) the worst point.

  In the fifth third case   ~({\it poset-like scale}  \(\Rightarrow\)
 {\it ordinal scale}),
 analogical methods
  (as for the transformation type 2)
 can be used, for example:~
%
  series detection of Pareto-layers, etc.
%

 For the case eight, Fig. 14 depicts the layers of quality
 (an ordinal scale \(D\) as in Fig. 6b):~
 (i) the ideal solution \(e(S^{I})\),
 (ii) Pareto-efficient points
 (i.e., \( \{e(S^{p}_{1}),e(S^{p}_{2}),e(S^{p}_{3}),e(S^{p}_{4})\}\),
 (iii) points of the next layer of quality,
 (i.e., \( \{e(S'_{1}),e(S'_{2}),e(S'_{3})\}\),
 (iv) another point (the next layer of quality)
 (i.e, \(e(S'')\), and
 (v) the worst point.
 Here, \(I\), \(S^{p}_{i} (i=\overline{1,4})\),
 \(S'_{j} (j=\overline{1,3})\),
 \(S''\) correspond to system versions.

 Analogically (case nine), the total ordinal scale for system quality is depicted
 for poset-like scale based on interval  multiset estimates
 in Fig. 9:
 (i) the ideal solution,
%
 (ii) Pareto-efficient points;
%
%
%
 (iii) points of the next layer of quality;
%
%
%
 (iv) another point (the next layer of quality);
 and
 (v) the worst point.

\begin{center}
\begin{picture}(60,49)

\put(018,00){\makebox(0,0)[bl]{Fig. 12. Ordinal scale
 \( \Rightarrow\)
 ordinal scale}}

\put(06.4,5){\makebox(0,0)[bl]{(a) mapping \(1\)}}

\put(05,48){\circle*{1.4}}

\put(07,48){\line(1,0){06}} \put(13,48){\line(1,-1){4}}
\put(17,44){\line(1,0){6}} \put(23,44){\vector(2,-1){5}}

\put(00,47){\makebox(0,0)[bl]{\(1\)}}

\put(05,48){\vector(0,-1){4}}

\put(00,42){\makebox(0,0)[bl]{\(2\)}}

\put(05,43){\circle*{1.4}} \put(05,43){\vector(0,-1){4}}

\put(07,43){\line(1,0){5}} \put(12,43){\line(2,-1){6}}
\put(18,40){\vector(1,0){10}}

\put(00,37){\makebox(0,0)[bl]{\(3\)}}

\put(05,38){\circle*{1.4}}

\put(07,38){\line(1,0){6}} \put(13,38){\line(2,-1){6}}
\put(19,35){\vector(1,0){09}}

\put(05,38){\vector(0,-1){4}}

\put(03.5,31){\makebox(0,0)[bl]{...}}


\put(05,29){\vector(0,-1){4}}


\put(00,23){\makebox(0,0)[bl]{\(\kappa'\)}}

\put(5,24){\circle*{1.4}}

\put(07,24){\line(1,0){6}} \put(13,24){\line(2,-1){6}}
\put(19,21){\vector(1,0){09}}

\put(05,23){\vector(0,-1){4}}

\put(03.5,17){\makebox(0,0)[bl]{...}}

\put(15,18){\makebox(0,0)[bl]{...}}


\put(05,15){\vector(0,-1){4}}


\put(00,09){\makebox(0,0)[bl]{\(\kappa''\)}}

\put(5,10){\circle*{1.4}}

\put(07,10){\line(1,0){05}} \put(12,10){\line(1,1){6}}
\put(18,16){\line(1,0){06}} \put(24,16){\vector(1,1){04}}


\put(14.5,29.5){\makebox(0,0)[bl]{. . .}}








\put(32,39.5){\makebox(0,0)[bl]{\(1\)}}

\put(30,40){\circle*{0.9}} \put(30,40){\circle{1.9}}

\put(30,40){\vector(0,-1){4}}

\put(32,34.5){\makebox(0,0)[bl]{\(2\)}}

\put(30,35){\circle*{0.9}} \put(30,35){\circle{1.9}}

\put(30,35){\vector(0,-1){4}}

\put(28.5,28){\makebox(0,0)[bl]{...}}


\put(30,26){\vector(0,-1){4}}


\put(32,20.5){\makebox(0,0)[bl]{\(\kappa'''\)}}

\put(30,21){\circle*{0.9}} \put(30,21){\circle{1.9}}


\end{picture}
%
\begin{picture}(40,49)

\put(06.4,5){\makebox(0,0)[bl]{(b) mapping \(2\)}}






\put(00,42){\makebox(0,0)[bl]{\(1\)}}

\put(05,43){\circle*{1.4}}

\put(07,43){\vector(1,0){21}}

\put(05,43){\vector(0,-1){4}}

\put(00,37){\makebox(0,0)[bl]{\(2\)}}

\put(05,38){\circle*{1.4}}

\put(07,38){\vector(1,0){21}}

\put(05,38){\vector(0,-1){4}}

\put(03.5,31){\makebox(0,0)[bl]{...}}


\put(05,29){\vector(0,-1){4}}


\put(00,23){\makebox(0,0)[bl]{\(\kappa'\)}}

\put(5,24){\circle*{1.4}}

\put(07,24){\vector(1,0){21}}

\put(05,23){\vector(0,-1){4}}

\put(03.5,17){\makebox(0,0)[bl]{...}}

\put(15,20){\makebox(0,0)[bl]{...}}


\put(05,15){\vector(0,-1){4}}


\put(00,09){\makebox(0,0)[bl]{\(\kappa''\)}}

\put(5,10){\circle*{1.4}}

\put(07,10){\line(1,0){04}} \put(11,10){\line(1,1){05}}

\put(16,15){\line(1,1){04}}

\put(20,19){\line(1,0){04}} \put(24,19){\vector(1,1){04}}


\put(14.5,29.5){\makebox(0,0)[bl]{. . .}}







\put(32,42.5){\makebox(0,0)[bl]{\(1\)}}

\put(30,43){\circle*{0.9}} \put(30,43){\circle{1.9}}

\put(30,43){\vector(0,-1){4}}

\put(32,37.5){\makebox(0,0)[bl]{\(2\)}}

\put(30,38){\circle*{0.9}} \put(30,38){\circle{1.9}}

\put(30,39){\vector(0,-1){4}}

\put(28.5,31){\makebox(0,0)[bl]{...}}


\put(30,29){\vector(0,-1){4}}


\put(32,23.5){\makebox(0,0)[bl]{\(\kappa'\)}}

\put(30,24){\circle*{0.9}} \put(30,24){\circle{1.9}}

\end{picture}
\end{center}

\begin{center}
\begin{picture}(62,45)

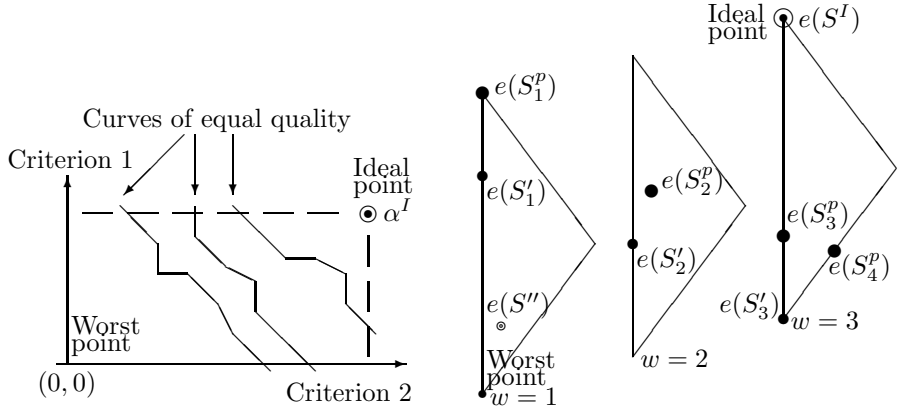
\put(02,00){\makebox(0,0)[bl]{Fig. 13. Curves of
 equal quality}}

\put(06.5,10){\line(1,0){3}}

\put(04,05.5){\makebox(0,0)[bl]{\((0,0)\)}}

\put(08,10){\vector(0,1){25}} \put(08,10){\vector(1,0){45}}

\put(37,05){\makebox(0,0)[bl]{Criterion 2}}
\put(00,36){\makebox(0,0)[bl]{Criterion 1}}


\put(10,40.8){\makebox(0,0)[bl]{Curves of equal quality}}

\put(23.5,40.5){\vector(-1,-1){8}} \put(25,40.5){\vector(0,-1){8}}
\put(30,40.5){\vector(0,-1){8}}





\put(10,30){\line(1,0){4}} \put(16,30){\line(1,0){4}}
\put(22,30){\line(1,0){4}} \put(28,30){\line(1,0){4}}
\put(34,30){\line(1,0){4}} \put(40,30){\line(1,0){4}}



\put(48,11){\line(0,1){4}} \put(48,17){\line(0,1){4}}
\put(48,23){\line(0,1){4}}


\put(30,31){\line(1,-1){7}} \put(37,24){\line(1,0){4}}
\put(41,24){\line(2,-1){4}} \put(45,22){\line(0,-1){4}}
\put(45,18){\line(1,-1){4}}


\put(25,31){\line(0,-1){4}} \put(25,27){\line(1,-1){4}}
\put(29,23){\line(2,-1){4}} \put(33,21){\line(0,-1){4}}
\put(33,17){\line(1,-1){8}}


\put(15,31){\line(1,-1){5}} \put(20,26){\line(0,-1){4}}
\put(20,22){\line(1,0){4}} \put(24,22){\line(1,-1){4}}
\put(28,18){\line(1,-2){2}} \put(30,14){\line(1,-1){5}}


\put(48,30){\circle*{1}}\put(48,30){\circle{2}}

\put(46,35){\makebox(0,0)[bl]{Ideal}}
\put(46,32){\makebox(0,0)[bl]{point}}
\put(50,29){\makebox(0,0)[bl]{\(\alpha^{I}\)}}






\put(08.5,14){\makebox(0,0)[bl]{Worst}}
\put(08.5,11){\makebox(0,0)[bl]{point}}

\end{picture}
%
\begin{picture}(55,59)

\put(0.5,00){\makebox(0,0)[bl]{Fig. 14. Layers at
 poset-like scale}}

\put(00,06){\line(0,1){40}} \put(00,06){\line(3,4){15}}
\put(00,046){\line(3,-4){15}}

\put(20,011){\line(0,1){40}} \put(20,011){\line(3,4){15}}
\put(20,051){\line(3,-4){15}}

\put(40,016){\line(0,1){40}} \put(40,016){\line(3,4){15}}
\put(40,056){\line(3,-4){15}}

\put(00,46){\circle*{1.6}}
\put(01.6,45){\makebox(0,0)[bl]{\(e(S^{p}_{1})\)}}

\put(22.5,33){\circle*{1.6}}
\put(23.8,32.5){\makebox(0,0)[bl]{\(e(S^{p}_{2})\)}}

\put(40,27){\circle*{1.6}}
\put(40.5,28){\makebox(0,0)[bl]{\(e(S^{p}_{3})\)}}

\put(46.8,25){\circle*{1.6}}
\put(46,21){\makebox(0,0)[bl]{\(e(S^{p}_{4})\)}}


\put(00,35){\circle*{1.3}}
\put(00.5,31){\makebox(0,0)[bl]{\(e(S'_{1})\)}}

\put(20,26){\circle*{1.3}}
\put(20.5,22){\makebox(0,0)[bl]{\(e(S'_{2})\)}}

\put(40,16){\circle*{1.3}}
\put(31.5,16){\makebox(0,0)[bl]{\(e(S'_{3})\)}}


\put(02.5,15){\circle{0.5}} \put(02.5,15){\circle{1.2}}

\put(00.5,16){\makebox(0,0)[bl]{\(e(S'')\)}}





\put(40,56){\circle*{1}} \put(40,56){\circle{2.5}}

\put(30,55.5){\makebox(0,0)[bl]{Ideal}}
\put(30,52.5){\makebox(0,0)[bl]{point}}

\put(42,54){\makebox(0,0)[bl]{\(e(S^{I})\)}}


\put(00.5,09.5){\makebox(0,0)[bl]{Worst}}
\put(00.5,06.5){\makebox(0,0)[bl]{point}}

\put(00,06){\circle*{0.9}}

\put(01,04.6){\makebox(0,0)[bl]{\(w=1\)}}
\put(21,09.6){\makebox(0,0)[bl]{\(w=2\)}}
\put(41,14.6){\makebox(0,0)[bl]{\(w=3\)}}

\end{picture}
\end{center}





\section{Integration of Scales and System Quality}

 Some  approaches to integration
 of system component/compatibility estimates into a total system estimate
(i.e., system evaluation) are
 the following (Table 3):~
 {\bf 1.}  {\it Quantitative estimates}  ~\(\Rightarrow\)
 {\it integrated quantitative estimate}:
 (1.1) utility function approaches
 (e.g., \cite{fis70,kee76}),
 (1.2) AHP and its modifications
 (e.g., \cite{saaty88}),
 (1.3) TOPSIS-like methods
 (TOPSIS: technique for order performance by similarity to ideal
 solution)
 (e.g., \cite{la94,shih07}),
%
%
 (1.4) frameworks based on
 analysis and usage of reference solutions,
 and
 (1.5) hybrid methods.

{\bf 2.}  {\it Quantitative  estimates and ordinal estimates}
\(\Rightarrow\)
 {\it integrated ordinal estimates}
 (or sorting problems)
 (e.g., \cite{lev98,lev06,zap02}):~
%
 (2.1) usage of ordinal scale \(D\)
  (e.g., \cite{lev98,lev06}),
 (2.2) series detection of Pareto-efficient points
 (as Pareto-layers)
  (e.g., \cite{mirkin79,pareto71}),
 (2.3) series detection of maximal points,
 (2.4) outranking techniques
  (e.g., \cite{brans84,roy96}),
%
%
 (2.5) frameworks based on
 analysis and usage of reference solutions,
 and
 (2.6) hybrid/composite methods
 (e.g., \cite{lev12b,levmih88}).

 {\bf 3.}  {\it Ordinal estimates}  \(\Rightarrow\)
 {\it integrated ordinal estimates}
(or sorting problems)
 (e.g., \cite{lev98,lev06,zap02}):~
%
 (3.1) integration tables
 (e.g., \cite{glo84,lev06}),
 (3.2) man-machine interactive procedures  (expert judgment)
 to design the class bounds at the total system quality domain
 (i.e., ordinal scale for system quality) (Fig. 15)
  (e.g., \cite{larmf86,levmih88}),
  (3.3) man-machine interactive procedures  (expert and logical methods)
  to design the class bounds at the total system quality domain
 (i.e., ordinal scale for system quality) (Fig. 15)
  (e.g., \cite{alek76,levmih88,ser83}),
 (3.4) frameworks based on
 analysis and usage of reference solutions,
 and
 (3.5) hybrid methods
 (e.g., \cite{lev12b,levmih88}).

 {\bf 4.}  {\it Ordinal estimates}  \(\Rightarrow\)
 {\it integrated poset-like estimate}
 (e.g., \cite{lev98,lev06,lev12morph}):~
%
  (4.1) computing the integrated poset-like estimates,
  (4.2) usage of expert judgment to get the
 integrated poset-like estimates.

 {\bf 5.}  {\it Poset-like estimates}  \(\Rightarrow\)
 {\it integrated poset-like estimate}
 (e.g., \cite{lev12a}):
%
 (5.1) integrated estimate,
 (5.2) median-like estimate,
 (5.3) usage of expert judgment.

 {\bf 6.}  {\it Vector-like  estimates}  \(\Rightarrow\)
 {\it integrated vector-like estimate}:~
%
 (6.1) unification of the initial multicriteria
 (i.e., multidimensional) domains,
 (6.2) simple integration of the initial multicriteria
 (i.e., multidimensional)
 domains (e.g., summarization by components),
 (6.3) special mappings.

\begin{center}
\begin{picture}(117,169)

\put(02,165){\makebox(0,0)[bl]{Table 3. Approaches to integration
of component/compatibility estimates}}

\put(00,0){\line(1,0){117}} \put(00,148){\line(1,0){117}}
\put(00,163){\line(1,0){117}}

\put(00,0){\line(0,1){163}} \put(27,0){\line(0,1){163}}
\put(49,0){\line(0,1){163}} \put(71,0){\line(0,1){163}}
\put(98,0){\line(0,1){163}} \put(117,0){\line(0,1){163}}


\put(01,158){\makebox(0,0)[bl]{Methods}}
\put(01,154){\makebox(0,0)[bl]{}}

\put(28,158){\makebox(0,0)[bl]{Scales for}}
\put(28,154){\makebox(0,0)[bl]{system}}
\put(28,150){\makebox(0,0)[bl]{components }}

\put(50,158){\makebox(0,0)[bl]{Scale for}}
\put(50,154){\makebox(0,0)[bl]{total system }}
\put(50,150){\makebox(0,0)[bl]{quality}}

\put(72,158){\makebox(0,0)[bl]{Type of}}
\put(72,154){\makebox(0,0)[bl]{integration}}

\put(99,158){\makebox(0,0)[bl]{Some }}
\put(99,154){\makebox(0,0)[bl]{sources}}


\put(0.5,143){\makebox(0,0)[bl]{1.Utility analysis,}}
\put(1.5,139){\makebox(0,0)[bl]{TOPSIS, AHP}}

\put(28,143){\makebox(0,0)[bl]{Quantitative}}
\put(28,139){\makebox(0,0)[bl]{}}

\put(50,143){\makebox(0,0)[bl]{Quantitative}}

\put(72,143){\makebox(0,0)[bl]{Utility function,}}
\put(72,139){\makebox(0,0)[bl]{TOPSIS, AHP}}

\put(99,143){\makebox(0,0)[bl]{\cite{fis70,kee76,la94}}}
\put(99,139){\makebox(0,0)[bl]{\cite{saaty88,shih07}}}


\put(0.5,133.5){\makebox(0,0)[bl]{2.Integration  }}
\put(1.5,130){\makebox(0,0)[bl]{tables}}

\put(28,134){\makebox(0,0)[bl]{Ordinal }}
\put(28,130){\makebox(0,0)[bl]{}}

\put(50,134){\makebox(0,0)[bl]{Ordinal}}

\put(72,134){\makebox(0,0)[bl]{Hierarchical}}
\put(72,129.5){\makebox(0,0)[bl]{integration}}
\put(72,126){\makebox(0,0)[bl]{tables}}

\put(99,134){\makebox(0,0)[bl]{\cite{glo84,lev06}}}


\put(0.5,121){\makebox(0,0)[bl]{3.Pareto-}}
\put(1.5,117){\makebox(0,0)[bl]{approach}}

\put(27.5,120.5){\makebox(0,0)[bl]{Quantitative, }}
\put(28,117){\makebox(0,0)[bl]{ordinal}}

\put(50,121){\makebox(0,0)[bl]{Ordinal}}

\put(72,121){\makebox(0,0)[bl]{Detection of}}
\put(72,117){\makebox(0,0)[bl]{Pareto-layer}}

\put(99,121){\makebox(0,0)[bl]{\cite{mirkin79,pareto71}}}


\put(0.5,111.5){\makebox(0,0)[bl]{4.Outranking }}
\put(1.5,108.5){\makebox(0,0)[bl]{methods}}
\put(1.5,104){\makebox(0,0)[bl]{(ELECTRE,}}
\put(1,100){\makebox(0,0)[bl]{PROMETHEE)}}

\put(27.5,111.5){\makebox(0,0)[bl]{Quantitative, }}
\put(28,108){\makebox(0,0)[bl]{ordinal}}

\put(50,112){\makebox(0,0)[bl]{Ordinal}}

\put(72,112){\makebox(0,0)[bl]{Detection of}}
\put(72,108){\makebox(0,0)[bl]{dominating}}
\put(72,104){\makebox(0,0)[bl]{points}}

\put(99,112){\makebox(0,0)[bl]{\cite{brans84,roy96}}}


\put(0.5,95){\makebox(0,0)[bl]{5.Layer of}}
\put(1.5,91){\makebox(0,0)[bl]{maximal}}
\put(1.5,86.5){\makebox(0,0)[bl]{(minimal)}}
\put(1.5,83){\makebox(0,0)[bl]{elements}}

\put(27.5,94.5){\makebox(0,0)[bl]{Quantitative, }}
\put(28,91){\makebox(0,0)[bl]{ordinal}}

\put(50,95){\makebox(0,0)[bl]{Ordinal}}

\put(72,95){\makebox(0,0)[bl]{Detection of}}
\put(72,90.5){\makebox(0,0)[bl]{maximal (or/}}
\put(72,86.5){\makebox(0,0)[bl]{and minimal)}}
\put(72,83){\makebox(0,0)[bl]{elements}}



\put(0.5,78){\makebox(0,0)[bl]{6.Man-machine }}
\put(1.5,74){\makebox(0,0)[bl]{procedure}}
\put(1.5,70){\makebox(0,0)[bl]{(expert}}
\put(1.5,66){\makebox(0,0)[bl]{judgment)}}

\put(28,78){\makebox(0,0)[bl]{Ordinal}}

\put(50,78){\makebox(0,0)[bl]{Ordinal}}

\put(72,78){\makebox(0,0)[bl]{Dividing class}}
\put(72,74){\makebox(0,0)[bl]{bounds for}}
\put(72,70){\makebox(0,0)[bl]{multicriteria}}
\put(72,66){\makebox(0,0)[bl]{domain}}

\put(99,78){\makebox(0,0)[bl]{\cite{larmf86}}}


\put(0.5,61){\makebox(0,0)[bl]{7.Interactive}}
\put(1.5,57){\makebox(0,0)[bl]{procedure}}
\put(1.5,53){\makebox(0,0)[bl]{(expert and}}
\put(1.5,49){\makebox(0,0)[bl]{logical methods)}}

\put(28,61){\makebox(0,0)[bl]{Ordinal}}

\put(50,61){\makebox(0,0)[bl]{Ordinal}}

\put(72,61){\makebox(0,0)[bl]{Dividing class}}
\put(72,57){\makebox(0,0)[bl]{bounds for}}
\put(72,53){\makebox(0,0)[bl]{multicriteria}}
\put(72,49){\makebox(0,0)[bl]{domain}}

\put(99,61){\makebox(0,0)[bl]{\cite{alek76,levmih88,ser83}}}


\put(0.5,44){\makebox(0,0)[bl]{8.Unification}}
\put(1.5,40){\makebox(0,0)[bl]{of}}
\put(1.5,36){\makebox(0,0)[bl]{measurement}}
\put(1.5,32){\makebox(0,0)[bl]{domains}}

\put(28,44){\makebox(0,0)[bl]{Multicriteria}}
\put(28,40){\makebox(0,0)[bl]{description}}

\put(50,44){\makebox(0,0)[bl]{Multicriteria}}
\put(50,40){\makebox(0,0)[bl]{description}}

\put(72,43.5){\makebox(0,0)[bl]{Integration }}
\put(72,40){\makebox(0,0)[bl]{of domains}}
\put(72,35.5){\makebox(0,0)[bl]{(unification,}}
\put(72,32){\makebox(0,0)[bl]{consensus)}}

\put(99,44){\makebox(0,0)[bl]{\cite{lev98,lev06,lev12a}}}


\put(0.5,27){\makebox(0,0)[bl]{9.HMMD}}
\put(1.5,23){\makebox(0,0)[bl]{with ordinal}}
\put(1.5,19){\makebox(0,0)[bl]{estimates}}

\put(28,27){\makebox(0,0)[bl]{Ordinal}}
\put(28,23){\makebox(0,0)[bl]{estimates}}

\put(50,27){\makebox(0,0)[bl]{Poset based}}
\put(50,23){\makebox(0,0)[bl]{on ordinal }}
\put(50,19){\makebox(0,0)[bl]{estimates}}

\put(72,27){\makebox(0,0)[bl]{Detection of }}
\put(72,23){\makebox(0,0)[bl]{Pareto layer}}
\put(72,19){\makebox(0,0)[bl]{}}

\put(99,27){\makebox(0,0)[bl]{\cite{lev98,lev06,lev09}}}
\put(99,23){\makebox(0,0)[bl]{\cite{lev12morph}}}


\put(0.5,14){\makebox(0,0)[bl]{10.HMMD}}
\put(1.5,10){\makebox(0,0)[bl]{with interval}}
\put(1.5,06){\makebox(0,0)[bl]{multiset}}
\put(1.5,02){\makebox(0,0)[bl]{estimates}}

\put(28,14){\makebox(0,0)[bl]{Interval }}
\put(28,10){\makebox(0,0)[bl]{multiset}}
\put(28,06){\makebox(0,0)[bl]{estimates}}

\put(50,14){\makebox(0,0)[bl]{Poset based}}
\put(50,10){\makebox(0,0)[bl]{on interval}}
\put(50,06){\makebox(0,0)[bl]{multiset}}
\put(50,02){\makebox(0,0)[bl]{estimates}}

\put(72,13.5){\makebox(0,0)[bl]{(a) Integrated}}
\put(72,10.5){\makebox(0,0)[bl]{or median-like}}
\put(72,06){\makebox(0,0)[bl]{estimate, }}
\put(72,02){\makebox(0,0)[bl]{(b) Pareto layer}}

\put(99,14){\makebox(0,0)[bl]{\cite{lev12a}}}

\end{picture}
\end{center}

 From the engineering viewpoint (i.e, experience of domain experts),
 it may be reasonable
 to illustrate two methods:
 (a) integrated tables
 (Fig. 16 and Fig. 17; numerical examples of system, integrated of tables, and system evaluation), and
 (b) TOPSIS (Fig. 18;  an illustration
  of an extended version for several ideal points).

\begin{center}
\begin{picture}(74,75)

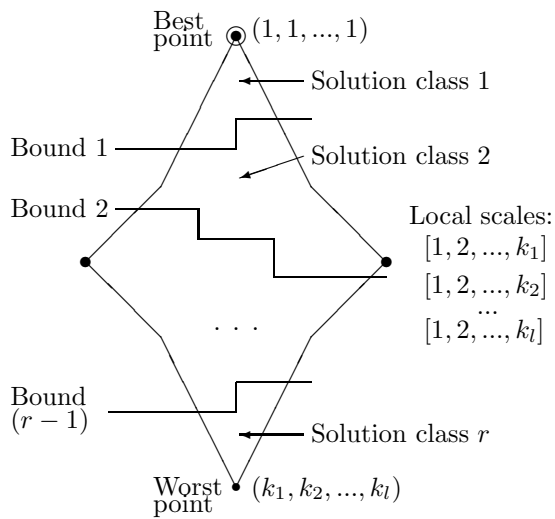
\put(00,00){\makebox(0,0)[bl]{Fig. 15. Class bounds
 for ordinal system quality}}

\put(19,71){\makebox(0,0)[bl]{Best}}
\put(19,68){\makebox(0,0)[bl]{point}}

\put(32,69){\makebox(0,0)[bl]{\((1,1,...,1)\)}}

\put(30,70){\circle*{1.5}} \put(30,70){\circle{2.4}}


\put(40,63){\makebox(0,0)[bl]{Solution class \(1\)}}
\put(39,64){\vector(-1,0){08.5}}


\put(40,53){\makebox(0,0)[bl]{Solution class \(2\)}}
\put(39,54){\vector(-3,-1){08.5}}


\put(53,45){\makebox(0,0)[bl]{Local scales:}}
\put(55,40){\makebox(0,0)[bl]{\([1,2,...,k_{1}]\)}}
\put(55,35){\makebox(0,0)[bl]{\([1,2,...,k_{2}]\)}}
\put(62,33){\makebox(0,0)[bl]{\(...\)}}
\put(55,29){\makebox(0,0)[bl]{\([1,2,...,k_{l}]\)}}


\put(20,50){\line(1,2){10}} \put(40,50){\line(-1,2){10}}

\put(10,40){\line(1,1){10}} \put(50,40){\line(-1,1){10}}

\put(10,40){\circle*{1.4}} \put(50,40){\circle*{1.4}}

\put(20,30){\line(-1,1){10}} \put(40,30){\line(1,1){10}}

\put(30,10){\line(-1,2){10}} \put(30,10){\line(1,2){10}}


\put(00,54){\makebox(0,0)[bl]{Bound \(1\)}}

\put(14,55){\line(1,0){16}} \put(30,55){\line(0,1){04}}
\put(30,59){\line(1,0){10}}

\put(00,46){\makebox(0,0)[bl]{Bound \(2\)}}

\put(14,47){\line(1,0){11}} \put(25,47){\line(0,-1){04}}
\put(25,43){\line(1,0){10}} \put(35,43){\line(0,-1){05}}
\put(35,38){\line(1,0){15}}

\put(27,31){\makebox(0,0)[bl]{. . .}}

\put(00,21){\makebox(0,0)[bl]{Bound}}
\put(00,17){\makebox(0,0)[bl]{(\(r-1\))}}

\put(13,20){\line(1,0){17}} \put(30,20){\line(0,1){04}}
\put(30,24){\line(1,0){10}}


\put(40,16){\makebox(0,0)[bl]{Solution class \(r\)}}
\put(39,17){\vector(-1,0){08.5}}


\put(30,10){\circle*{0.9}}

\put(19,09){\makebox(0,0)[bl]{Worst}}
\put(19,06){\makebox(0,0)[bl]{point}}

\put(32,08){\makebox(0,0)[bl]{\((k_{1},k_{2},...,k_{l})\)}}

\end{picture}
\end{center}

\begin{center}
\begin{picture}(116,36)

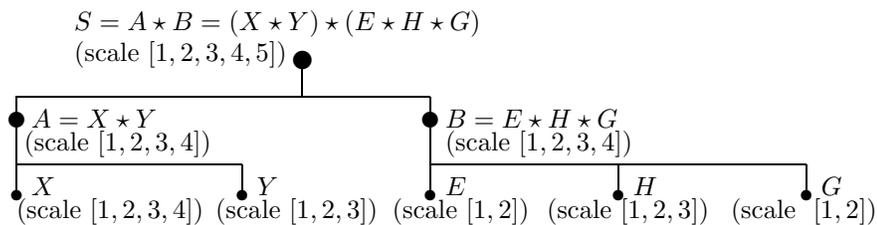
\put(08,02){\makebox(0,0)[bl]{Fig. 16. Example of system
structure, scales for components}}


\put(00,20){\circle*{2}} \put(02,19){\makebox(0,8)[bl]{\(A = X
\star Y\)}}

\put(01,15){\makebox(0,8)[bl]{(scale \([1,2,3,4]\))}}

\put(00,20){\line(0,-1){6}}

\put(00,14){\line(1,0){30}}


\put(00,14){\line(0,-1){4}}

\put(00,10){\circle*{1.3}}

\put(02,10){\makebox(0,8)[bl]{\(X\)}}
\put(00,06){\makebox(0,8)[bl]{(scale \([1,2,3,4]\))}}


\put(30,14){\line(0,-1){4}}

\put(30,10){\circle*{1.3}}

\put(32,10){\makebox(0,8)[bl]{\(Y\)}}
\put(26.5,06){\makebox(0,8)[bl]{(scale \([1,2,3]\))}}


\put(55,20){\circle*{2}} \put(57,19){\makebox(0,8)[bl]{\(B = E
\star H \star G\)}}

\put(57,15){\makebox(0,8)[bl]{(scale \([1,2,3,4]\))}}

\put(55,14){\line(0,1){6}}

\put(55,14){\line(1,0){50}}


\put(55,14){\line(0,-1){4}}

\put(55,10){\circle*{1.3}}

\put(57,10){\makebox(0,8)[bl]{\(E\)}}
\put(50,06){\makebox(0,8)[bl]{(scale \([1,2]\))}}


\put(80,14){\line(0,-1){4}}

\put(80,10){\circle*{1.3}}

\put(82,10){\makebox(0,8)[bl]{\(H\)}}
\put(70.5,06){\makebox(0,8)[bl]{(scale \([1,2,3]\))}}


\put(105,14){\line(0,-1){4}}

\put(105,10){\circle*{1.3}}

\put(107,10){\makebox(0,8)[bl]{\(G\)}}
\put(95,06){\makebox(0,8)[bl]{(scale ~\([1,2]\))}}


\put(00,19){\line(0,1){4}} \put(55,19){\line(0,1){4}}

\put(00,23){\line(1,0){55}}


\put(38,23){\line(0,1){05}} \put(38,28){\circle*{2.6}}

\put(07.7,31){\makebox(0,8)[bl]{\(S=A\star B = (X \star Y )\star
 (E \star H \star G)\)}}
\put(07.7,27){\makebox(0,8)[bl]{(scale \([1,2,3,4,5]\))}}

\end{picture}
\end{center}

\begin{center}
\begin{picture}(105,102)
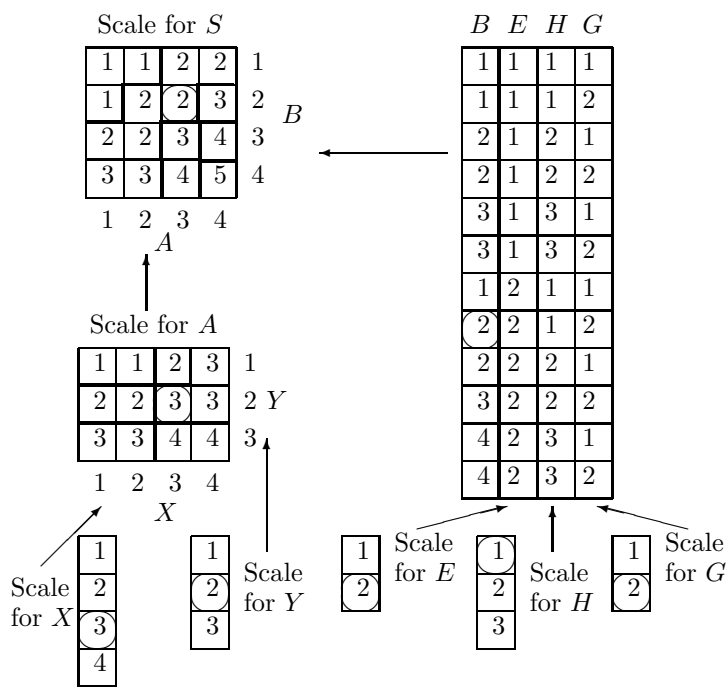
\put(18,02){\makebox(0,0)[bl]{Fig. 17. Integration of scales by
tables}}


\put(09,10){\line(0,1){20}} \put(14,10){\line(0,1){20}}
\put(09,10){\line(1,0){05}}

\put(09,15){\line(1,0){05}} \put(09,20){\line(1,0){05}}
\put(09,25){\line(1,0){05}} \put(09,30){\line(1,0){05}}

\put(11,27){\makebox(0,8)[bl]{\(1\)}}
\put(11,22){\makebox(0,8)[bl]{\(2\)}}
\put(11,17){\makebox(0,8)[bl]{\(3\)}}

\put(11.5,17.5){\oval(5,5)}

\put(11,12){\makebox(0,8)[bl]{\(4\)}}

\put(00,22){\makebox(0,8)[bl]{Scale}}
\put(00,18){\makebox(0,8)[bl]{for \(X\)}}

\put(04,26){\vector(1,1){8}}


\put(24,15){\line(0,1){15}} \put(29,15){\line(0,1){15}}
\put(24,15){\line(1,0){05}} \put(24,20){\line(1,0){05}}
\put(24,25){\line(1,0){05}} \put(24,30){\line(1,0){05}}

\put(26,27){\makebox(0,8)[bl]{\(1\)}}
\put(26,22){\makebox(0,8)[bl]{\(2\)}}

\put(26.5,22.5){\oval(5,5)}

\put(26,17){\makebox(0,8)[bl]{\(3\)}}

\put(31,24){\makebox(0,8)[bl]{Scale}}
\put(31,20){\makebox(0,8)[bl]{for \(Y\)}}

\put(34,28){\vector(0,1){15}}


\put(18,60){\vector(0,1){7.5}}

\put(10.5,57){\makebox(0,8)[bl]{Scale for \(A\)}}

\put(34,47){\makebox(0,8)[bl]{\(Y\)}}

\put(31,52){\makebox(0,8)[bl]{\(1\)}}
\put(31,47){\makebox(0,8)[bl]{\(2\)}}
\put(31,42){\makebox(0,8)[bl]{\(3\)}}

\put(09,40){\line(0,1){15}} \put(14,40){\line(0,1){15}}
\put(19,40){\line(0,1){15}} \put(19.3,40){\line(0,1){15}}
\put(24,40){\line(0,1){15}} \put(23.8,50.2){\line(0,1){4.8}}
\put(29,40){\line(0,1){15}}

\put(09,40){\line(1,0){20}} \put(09,44.9){\line(1,0){20}}
\put(09,45.1){\line(1,0){20}} \put(09,50.2){\line(1,0){14.8}}
\put(09,50){\line(1,0){20}} \put(09,55){\line(1,0){20}}

\put(11,36){\makebox(0,8)[bl]{\(1\)}}
\put(16,36){\makebox(0,8)[bl]{\(2\)}}
\put(21,36){\makebox(0,8)[bl]{\(3\)}}
\put(26,36){\makebox(0,8)[bl]{\(4\)}}

\put(19,32){\makebox(0,8)[bl]{\(X\)}}

\put(11,52){\makebox(0,8)[bl]{\(1\)}}
\put(11,47){\makebox(0,8)[bl]{\(2\)}}
\put(11,42){\makebox(0,8)[bl]{\(3\)}}

\put(16,52){\makebox(0,8)[bl]{\(1\)}}
\put(16,47){\makebox(0,8)[bl]{\(2\)}}
\put(16,42){\makebox(0,8)[bl]{\(3\)}}

\put(21,52){\makebox(0,8)[bl]{\(2\)}}
\put(21,47){\makebox(0,8)[bl]{\(3\)}}

\put(21.5,47.5){\oval(5,5)}

\put(21,42){\makebox(0,8)[bl]{\(4\)}}

\put(26,52){\makebox(0,8)[bl]{\(3\)}}
\put(26,47){\makebox(0,8)[bl]{\(3\)}}
\put(26,42){\makebox(0,8)[bl]{\(4\)}}


\put(11.5,97){\makebox(0,8)[bl]{Scale for \(S\)}}

\put(36,85){\makebox(0,8)[bl]{\(B\)}}

\put(32,92){\makebox(0,8)[bl]{\(1\)}}
\put(32,87){\makebox(0,8)[bl]{\(2\)}}
\put(32,82){\makebox(0,8)[bl]{\(3\)}}
\put(32,77){\makebox(0,8)[bl]{\(4\)}}

\put(10,75){\line(0,1){20}} \put(15,75){\line(0,1){20}}
\put(20,75){\line(0,1){20}} \put(25,75){\line(0,1){20}}
\put(30,75){\line(0,1){20}}

\put(10,75){\line(1,0){20}} \put(10,80){\line(1,0){20}}
\put(10,85){\line(1,0){20}} \put(10,90){\line(1,0){20}}
\put(10,95){\line(1,0){20}}

\put(25.2,75){\line(0,1){4.8}} \put(25.2,79.8){\line(1,0){4.8}}

\put(10,80.2){\line(1,0){9.8}} \put(19.8,80.2){\line(0,1){5}}
\put(19.8,85.2){\line(1,0){5}} \put(24.8,85.2){\line(0,1){5}}
\put(24.8,90.2){\line(1,0){4.8}}

\put(10,85.2){\line(1,0){4.8}} \put(14.8,85.2){\line(0,1){5}}
\put(14.8,90.2){\line(1,0){5}} \put(19.8,90.2){\line(0,1){4.8}}

\put(20.2,75){\line(0,1){5}} \put(20.2,80.2){\line(1,0){5}}
\put(25.2,80.2){\line(0,1){5}} \put(25.2,85.2){\line(1,0){4.8}}

\put(12,92){\makebox(0,8)[bl]{\(1\)}}
\put(12,87){\makebox(0,8)[bl]{\(1\)}}
\put(12,82){\makebox(0,8)[bl]{\(2\)}}
\put(12,77){\makebox(0,8)[bl]{\(3\)}}

\put(17,92){\makebox(0,8)[bl]{\(1\)}}
\put(17,87){\makebox(0,8)[bl]{\(2\)}}
\put(17,82){\makebox(0,8)[bl]{\(2\)}}
\put(17,77){\makebox(0,8)[bl]{\(3\)}}

\put(22,92){\makebox(0,8)[bl]{\(2\)}}
\put(22,87){\makebox(0,8)[bl]{\(2\)}}

\put(22.5,87.5){\oval(5,5)}

\put(22,82){\makebox(0,8)[bl]{\(3\)}}
\put(22,77){\makebox(0,8)[bl]{\(4\)}}

\put(27,92){\makebox(0,8)[bl]{\(2\)}}
\put(27,87){\makebox(0,8)[bl]{\(3\)}}
\put(27,82){\makebox(0,8)[bl]{\(4\)}}
\put(27,77){\makebox(0,8)[bl]{\(5\)}}

\put(12,71){\makebox(0,8)[bl]{\(1\)}}
\put(17,71){\makebox(0,8)[bl]{\(2\)}}
\put(22,71){\makebox(0,8)[bl]{\(3\)}}
\put(27,71){\makebox(0,8)[bl]{\(4\)}}

\put(19,68){\makebox(0,8)[bl]{\(A\)}}



\put(44,20){\line(0,1){10}} \put(49,20){\line(0,1){10}}


 \put(44,20){\line(1,0){05}}
\put(44,25){\line(1,0){05}} \put(44,30){\line(1,0){05}}

\put(46,27){\makebox(0,8)[bl]{\(1\)}}
\put(46,22){\makebox(0,8)[bl]{\(2\)}}

\put(46.5,22.5){\oval(5,5)}


\put(51,28){\makebox(0,8)[bl]{Scale}}
\put(51,24){\makebox(0,8)[bl]{for \(E\)}}

\put(54,31){\vector(4,1){12}}


\put(62,15){\line(0,1){15}} \put(67,15){\line(0,1){15}}
\put(62,15){\line(1,0){05}} \put(62,20){\line(1,0){05}}
\put(62,25){\line(1,0){05}} \put(62,30){\line(1,0){05}}

\put(64,27){\makebox(0,8)[bl]{\(1\)}}

\put(64.5,27.5){\oval(5,5)}

\put(64,22){\makebox(0,8)[bl]{\(2\)}}
\put(64,17){\makebox(0,8)[bl]{\(3\)}}

\put(69,24){\makebox(0,8)[bl]{Scale}}
\put(69,20){\makebox(0,8)[bl]{for \(H\)}}

\put(72,28){\vector(0,1){06}}


\put(80,20){\line(0,1){10}} \put(85,20){\line(0,1){10}}


 \put(80,20){\line(1,0){05}}
\put(80,25){\line(1,0){05}} \put(80,30){\line(1,0){05}}

\put(82,27){\makebox(0,8)[bl]{\(1\)}}
\put(82,22){\makebox(0,8)[bl]{\(2\)}}

\put(82.5,22.5){\oval(5,5)}


\put(87,28){\makebox(0,8)[bl]{Scale}}
\put(87,24){\makebox(0,8)[bl]{for \(G\)}}

\put(90,31){\vector(-4,1){12}}


\put(58,81){\vector(-1,0){17}}

\put(60,35){\line(0,1){60}} \put(65,35){\line(0,1){60}}
\put(70,35){\line(0,1){60}} \put(75,35){\line(0,1){60}}
\put(80,35){\line(0,1){60}}

\put(64.8,35){\line(0,1){60}} \put(64.9,35){\line(0,1){60}}

\put(60,35){\line(1,0){20}} \put(60,40){\line(1,0){20}}
\put(60,45){\line(1,0){20}} \put(60,50){\line(1,0){20}}
\put(60,55){\line(1,0){20}} \put(60,60){\line(1,0){20}}
\put(60,65){\line(1,0){20}} \put(60,70){\line(1,0){20}}
\put(60,75){\line(1,0){20}} \put(60,80){\line(1,0){20}}
\put(60,85){\line(1,0){20}} \put(60,90){\line(1,0){20}}
\put(60,95){\line(1,0){20}}

\put(61,97){\makebox(0,8)[bl]{\(B\)}}
\put(66,97){\makebox(0,8)[bl]{\(E\)}}
\put(71,97){\makebox(0,8)[bl]{\(H\)}}
\put(76,97){\makebox(0,8)[bl]{\(G\)}}


\put(62,92){\makebox(0,8)[bl]{\(1\)}}
\put(62,87){\makebox(0,8)[bl]{\(1\)}}
\put(62,82){\makebox(0,8)[bl]{\(2\)}}
\put(62,77){\makebox(0,8)[bl]{\(2\)}}
\put(62,72){\makebox(0,8)[bl]{\(3\)}}
\put(62,67){\makebox(0,8)[bl]{\(3\)}}
\put(62,62){\makebox(0,8)[bl]{\(1\)}}
\put(62,57){\makebox(0,8)[bl]{\(2\)}}

\put(62.5,57.5){\oval(5,5)}

\put(62,52){\makebox(0,8)[bl]{\(2\)}}
\put(62,47){\makebox(0,8)[bl]{\(3\)}}
\put(62,42){\makebox(0,8)[bl]{\(4\)}}
\put(62,37){\makebox(0,8)[bl]{\(4\)}}


\put(66,92){\makebox(0,8)[bl]{\(1\)}}
\put(66,87){\makebox(0,8)[bl]{\(1\)}}
\put(66,82){\makebox(0,8)[bl]{\(1\)}}
\put(66,77){\makebox(0,8)[bl]{\(1\)}}
\put(66,72){\makebox(0,8)[bl]{\(1\)}}
\put(66,67){\makebox(0,8)[bl]{\(1\)}}

\put(66,62){\makebox(0,8)[bl]{\(2\)}}
\put(66,57){\makebox(0,8)[bl]{\(2\)}}
\put(66,52){\makebox(0,8)[bl]{\(2\)}}
\put(66,47){\makebox(0,8)[bl]{\(2\)}}
\put(66,42){\makebox(0,8)[bl]{\(2\)}}
\put(66,37){\makebox(0,8)[bl]{\(2\)}}


\put(71,92){\makebox(0,8)[bl]{\(1\)}}
\put(71,87){\makebox(0,8)[bl]{\(1\)}}
\put(71,82){\makebox(0,8)[bl]{\(2\)}}
\put(71,77){\makebox(0,8)[bl]{\(2\)}}
\put(71,72){\makebox(0,8)[bl]{\(3\)}}
\put(71,67){\makebox(0,8)[bl]{\(3\)}}
\put(71,62){\makebox(0,8)[bl]{\(1\)}}
\put(71,57){\makebox(0,8)[bl]{\(1\)}}
\put(71,52){\makebox(0,8)[bl]{\(2\)}}
\put(71,47){\makebox(0,8)[bl]{\(2\)}}
\put(71,42){\makebox(0,8)[bl]{\(3\)}}
\put(71,37){\makebox(0,8)[bl]{\(3\)}}


\put(76,92){\makebox(0,8)[bl]{\(1\)}}
\put(76,87){\makebox(0,8)[bl]{\(2\)}}
\put(76,82){\makebox(0,8)[bl]{\(1\)}}

\put(76,77){\makebox(0,8)[bl]{\(2\)}}
\put(76,72){\makebox(0,8)[bl]{\(1\)}}
\put(76,67){\makebox(0,8)[bl]{\(2\)}}

\put(76,62){\makebox(0,8)[bl]{\(1\)}}
\put(76,57){\makebox(0,8)[bl]{\(2\)}}
\put(76,52){\makebox(0,8)[bl]{\(1\)}}

\put(76,47){\makebox(0,8)[bl]{\(2\)}}
\put(76,42){\makebox(0,8)[bl]{\(1\)}}
\put(76,37){\makebox(0,8)[bl]{\(2\)}}

\end{picture}
\end{center}

\begin{center}
\begin{picture}(82,46)

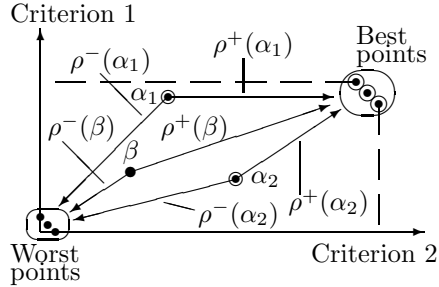
\put(00,00){\makebox(0,0)[bl]{Fig. 18. Illustration for
 TOPSIS-like methods}}


\put(07,09){\makebox(0,0)[bl]{Worst}}
\put(07,06){\makebox(0,0)[bl]{points}}

\put(12,14){\oval(5.6,4)}

\put(11,15){\circle*{0.9}} \put(12,14){\circle*{0.9}}
\put(13,13){\circle*{0.9}}


\put(11,13){\vector(0,1){27}} \put(11,13){\vector(1,0){51}}

\put(47,09){\makebox(0,0)[bl]{Criterion 2}}
\put(07,41){\makebox(0,0)[bl]{Criterion 1}}


\put(13,33){\line(1,0){4}} \put(19,33){\line(1,0){4}}
\put(25,33){\line(1,0){4}} \put(31,33){\line(1,0){4}}
\put(37,33){\line(1,0){4}} \put(43,33){\line(1,0){4}}
\put(49,33){\line(1,0){4}}


\put(56,14){\line(0,1){4}} \put(56,20){\line(0,1){4}}
\put(56,26){\line(0,1){4}}


\put(54.5,31.5){\oval(7.2,6)}

\put(53,33){\circle*{1.2}} \put(53,33){\circle{2}}
\put(54.5,31.5){\circle*{1.2}} \put(54.5,31.5){\circle{2}}
\put(56,30){\circle*{1.2}} \put(56,30){\circle{2}}

\put(53,38){\makebox(0,0)[bl]{Best}}
\put(53,35){\makebox(0,0)[bl]{points}}


\put(19,34){\line(1,-1){5.5}}
\put(15,34){\makebox(0,0)[bl]{\(\rho^{-}(\alpha_{1})\)}}

\put(38,36){\line(0,-1){4.5}}
\put(34,36){\makebox(0,0)[bl]{\(\rho^{+}(\alpha_{1})\)}}

\put(28,31){\vector(1,0){22}} \put(28,31){\vector(-1,-1){14.5}}

\put(28,31){\circle*{1.1}} \put(28,31){\circle{1.9}}
\put(23,30){\makebox(0,0)[bl]{\(\alpha_{1}\)}}

\put(37,20){\vector(3,2){14}} \put(37,20){\vector(-4,-1){21.6}}

\put(37,20){\circle*{1.1}} \put(37,20){\circle{1.9}}
\put(39,19){\makebox(0,0)[bl]{\(\alpha_{2}\)}}

\put(45.5,18){\line(0,1){7}}
\put(44,15){\makebox(0,0)[bl]{\(\rho^{+}(\alpha_{2})\)}}

\put(31.5,16){\line(-4,1){4}}
\put(32,13.5){\makebox(0,0)[bl]{\(\rho^{-}(\alpha_{2})\)}}


\put(12,25){\makebox(0,0)[bl]{\(\rho^{-}(\beta)\)}}
\put(14,25){\line(1,-1){5.5}}

\put(27,25){\makebox(0,0)[bl]{\(\rho^{+}(\beta)\)}}

\put(23,21){\vector(3,1){26.5}} \put(23,21){\vector(-3,-2){8}}

\put(23,21){\circle*{1.4}}
\put(22,22){\makebox(0,0)[bl]{\(\beta\)}}

\end{picture}
\end{center}

 In the basic versions of TOPSIS-like methods,
 transformation of multicriteria description
 of alternatives
  into a final ordinal
 scale is based on a simplification of the problem
 by consideration of proximity of the alternatives
 to the best solution.
%
 Generally, the alternatives are ordered
 by the vector \(\rho = (\rho^{-},\rho^{+} ) \)
 where
 \(\rho^{+} \) corresponds to proximity to
 the best point(s) (e.g., the ideal point(s)),
 \(\rho^{-} \) corresponds to proximity to
 the worst point(s).


%

\section{Numerical Examples}

 Here,
 simple numerical examples for four-component student team
 is described (Fig. 19) (system component compatibility is not examined).
 Table 4 contains initial estimates of
 team elements (i.e., alternatives for system components DAs)
 for four types of scales:

 (i) quantitative estimates (scale \((1,3)\), \(1\) corresponds to the best level);

 (ii) vector-like (two-element) ordinal estimates
 (scale \((x,y)\), \((1,1)\) corresponds to the best level,
 e.g., \(x\) corresponds to ``Mathematics'',
 \(y\) corresponds to ``Physics'');

 (iii) ordinal estimates (scale \([1,2,3]\),
 \(1\) corresponds to the best level);
 and

 (iv) interval multiset estimates
 (assessment problem \(P^{3,4}\), Fig. 7).

\begin{center}
\begin{picture}(84,49)
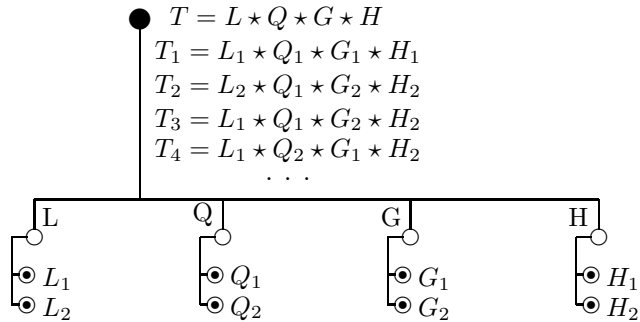
\put(07.7,00){\makebox(0,0)[bl] {Fig. 19. Example of
 four-component team}}

\put(4,05){\makebox(0,8)[bl]{\(L_{2}\)}}
\put(4,09){\makebox(0,8)[bl]{\(L_{1}\)}}

\put(29,05){\makebox(0,8)[bl]{\(Q_{2}\)}}
\put(29,09){\makebox(0,8)[bl]{\(Q_{1}\)}}

\put(54,05){\makebox(0,8)[bl]{\(G_{2}\)}}
\put(54,09){\makebox(0,8)[bl]{\(G_{1}\)}}

\put(79,05){\makebox(0,8)[bl]{\(H_{2}\)}}
\put(79,09){\makebox(0,8)[bl]{\(H_{1}\)}}

\put(03,16){\circle{2}} \put(28,16){\circle{2}}
\put(53,16){\circle{2}} \put(78,16){\circle{2}}

\put(00,16){\line(1,0){02}} \put(25,16){\line(1,0){02}}
\put(50,16){\line(1,0){02}} \put(75,16){\line(1,0){02}}

\put(00,16){\line(0,-1){09}} \put(25,16){\line(0,-1){09}}
\put(50,16){\line(0,-1){09}} \put(75,16){\line(0,-1){09}}

\put(75,07){\line(1,0){01}} \put(75,11){\line(1,0){01}}

\put(77,11){\circle{2}} \put(77,11){\circle*{1}}
\put(77,07){\circle{2}} \put(77,07){\circle*{1}}

\put(50,11){\line(1,0){01}} \put(50,07){\line(1,0){01}}

\put(52,11){\circle{2}} \put(52,11){\circle*{1}}
\put(52,07){\circle{2}} \put(52,07){\circle*{1}}

\put(25,07){\line(1,0){01}} \put(25,11){\line(1,0){01}}

\put(27,11){\circle{2}} \put(27,11){\circle*{1}}
\put(27,07){\circle{2}} \put(27,07){\circle*{1}}

\put(0,07){\line(1,0){01}} \put(0,11){\line(1,0){01}}

\put(2,07){\circle{2}} \put(2,11){\circle{2}}
\put(2,07){\circle*{1}} \put(2,11){\circle*{1}}

\put(03,21){\line(0,-1){04}} \put(28,21){\line(0,-1){04}}
\put(53,21){\line(0,-1){04}} \put(78,21){\line(0,-1){04}}

\put(03,21){\line(1,0){75}} \put(17,21){\line(0,1){23}}


\put(17,45){\circle*{2.7}}

\put(04,17.5){\makebox(0,8)[bl]{L}}
\put(24,17.5){\makebox(0,8)[bl]{Q}}
\put(49,17.5){\makebox(0,8)[bl]{G}}
\put(74,17.5){\makebox(0,8)[bl]{H}}

\put(21,43.5){\makebox(0,8)[bl]{\(T = L \star Q \star G \star
 H\)}}

\put(19,39){\makebox(0,8)[bl] {\(T_{1}=L_{1}\star Q_{1} \star
 G_{1} \star H_{1}\)}}

\put(19,34.5){\makebox(0,8)[bl] {\(T_{2}=L_{2}\star Q_{1} \star
 G_{2} \star H_{2}\)}}

\put(19,30){\makebox(0,8)[bl] {\(T_{3}=L_{1}\star Q_{1} \star
 G_{2} \star H_{2}\)}}

\put(19,26){\makebox(0,8)[bl] {\(T_{4}=L_{1}\star Q_{2} \star
 G_{1} \star H_{2}\)}}

\put(32.5,23.6){\makebox(0,8)[bl] { . . .}}

\end{picture}
\end{center}

\begin{center}
\begin{picture}(106,54)
\put(36,50){\makebox(0,0)[bl]{Table 4. Initial data}}

\put(00,00){\line(1,0){106}} \put(00,34){\line(1,0){106}}
\put(00,48){\line(1,0){106}}

\put(00,00){\line(0,1){48}} \put(07,00){\line(0,1){48}}
\put(28,00){\line(0,1){48}} \put(47,00){\line(0,1){48}}
\put(70,00){\line(0,1){48}} \put(106,00){\line(0,1){48}}


\put(01,44){\makebox(0,8)[bl]{DA}}

\put(7.7,43.5){\makebox(0,8)[bl]{Quantitative }}
\put(8,40){\makebox(0,8)[bl]{estimates}}
\put(8,36){\makebox(0,8)[bl]{(scale \((1,3)\))}}

\put(29,44){\makebox(0,8)[bl]{Vector-like}}
\put(29,40.5){\makebox(0,8)[bl]{estimates}}
\put(33,36){\makebox(0,8)[bl]{(\(x,y\))}}

\put(48,44){\makebox(0,8)[bl]{Ordinal}}
\put(48,40.5){\makebox(0,8)[bl]{estimates}}
\put(48,36){\makebox(0,8)[bl]{(scale \([1,2,3]\))}}

\put(71,44){\makebox(0,8)[bl]{Interval multiset}}
\put(71,40){\makebox(0,8)[bl]{estimates (assessment}}
\put(71,36){\makebox(0,8)[bl]{problem \(P^{3,4}\))}}


\put(01,29.5){\makebox(0,8)[bl]{\(L_{1}\)}}
\put(01,25.5){\makebox(0,8)[bl]{\(L_{2}\)}}
\put(01,21.5){\makebox(0,8)[bl]{\(Q_{1}\)}}
\put(01,17.5){\makebox(0,8)[bl]{\(Q_{2}\)}}
\put(01,13.5){\makebox(0,8)[bl]{\(G_{1}\)}}
\put(01,09.5){\makebox(0,8)[bl]{\(G_{2}\)}}
\put(01,05.5){\makebox(0,8)[bl]{\(H_{1}\)}}
\put(01,01.5){\makebox(0,8)[bl]{\(H_{2}\)}}


\put(15,30){\makebox(0,8)[bl]{\(1.5\)}}
\put(33.5,29.5){\makebox(0,8)[bl]{\((2,1)\)}}
\put(57.5,30){\makebox(0,8)[bl]{\(1\)}}
\put(83,29.5){\makebox(0,8)[bl]{\((3,1,0)\)}}


\put(15,26){\makebox(0,8)[bl]{\(1.8\)}}
\put(33.5,25.5){\makebox(0,8)[bl]{\((2,2)\)}}
\put(57.5,26){\makebox(0,8)[bl]{\(2\)}}
\put(83,25.5){\makebox(0,8)[bl]{\((0,4,0)\)}}


\put(15,22){\makebox(0,8)[bl]{\(1.1\)}}
\put(33.5,21.5){\makebox(0,8)[bl]{\((1,1)\)}}
\put(57.5,22){\makebox(0,8)[bl]{\(1\)}}
\put(83,21.5){\makebox(0,8)[bl]{\((4,0,0)\)}}


\put(15,18){\makebox(0,8)[bl]{\(2.7\)}}
\put(33.5,17.5){\makebox(0,8)[bl]{\((2,3)\)}}
\put(57.5,18){\makebox(0,8)[bl]{\(3\)}}
\put(83,17.5){\makebox(0,8)[bl]{\((0,3,1)\)}}


\put(15,14){\makebox(0,8)[bl]{\(1.2\)}}
\put(33.5,13.5){\makebox(0,8)[bl]{\((1,1)\)}}
\put(57.5,14){\makebox(0,8)[bl]{\(1\)}}
\put(83,13.5){\makebox(0,8)[bl]{\((3,1,0)\)}}


\put(15,10){\makebox(0,8)[bl]{\(2.4\)}}
\put(33.5,09.5){\makebox(0,8)[bl]{\((3,2)\)}}
\put(57.5,10){\makebox(0,8)[bl]{\(2\)}}
\put(83,09.5){\makebox(0,8)[bl]{\((1,2,1)\)}}


\put(15,06){\makebox(0,8)[bl]{\(1.4\)}}
\put(33.5,05.5){\makebox(0,8)[bl]{\((1,2)\)}}
\put(57.5,06){\makebox(0,8)[bl]{\(1\)}}
\put(83,05.5){\makebox(0,8)[bl]{\((2,2,0)\)}}


\put(15,02){\makebox(0,8)[bl]{\(3.1\)}}
\put(33.5,01.5){\makebox(0,8)[bl]{\((3,3)\)}}
\put(57.5,02){\makebox(0,8)[bl]{\(3\)}}
\put(83,01.5){\makebox(0,8)[bl]{\((0,2,2)\)}}


\end{picture}
\end{center}

 The following numerical examples are presented:

 {\bf Example 1.}
  Quantitative estimates of DAs are integrated
  by the simplest additive (i.e., utility) function
 (Fig. 20):¨
  \(f(T_{1}) = 1.5 + 1.1 + 1.2 + 1.4 = 5.2\) (the best solution),
  \(f(T_{2}) = 1.8 + 1.1 + 2.4 + 3.1 = 8.4\),
 \(f(T_{3}) = 1.5 + 1.1 + 2.4 + 3.1 = 8.1\),  and
 \(f(T_{4}) = 1.5 + 2.7 + 1.2 + 3.1 = 8.5\);
 the corresponding preference relation is:~
  \(T_{1} \succ  T_{3} \succ T_{2} \succ T_{4} \).

\begin{center}
\begin{picture}(100,22)

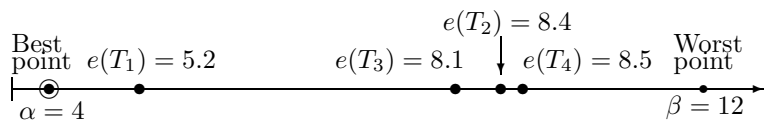
\put(03,00){\makebox(0,0)[bl]{Fig. 20. Resultant quantitative
 scale for modular solutions}}

\put(00,08.5){\line(0,1){3}}

\put(00,10){\vector(1,0){100}}


\put(01,06){\makebox(0,0)[bl]{\(\alpha=4\)}}

\put(05,10){\circle*{1.4}} \put(05,10){\circle{2.4}}

\put(00,15){\makebox(0,0)[bl]{Best}}
\put(00,12){\makebox(0,0)[bl]{point}}


\put(10,12){\makebox(0,0)[bl]{\(e(T_{1}) = 5.2\)}}

\put(17,10){\circle*{1.5}}


\put(43,12){\makebox(0,0)[bl]{\(e(T_{3}) = 8.1\)}}

\put(59,10){\circle*{1.5}}


\put(57.3,17){\makebox(0,0)[bl]{\(e(T_{2}) = 8.4\)}}

\put(65,17){\vector(0,-1){05}}

\put(65,10){\circle*{1.5}}


\put(68,12){\makebox(0,0)[bl]{\(e(T_{4}) = 8.5\)}}

\put(68,10){\circle*{1.5}}


\put(87,06){\makebox(0,0)[bl]{\(\beta = 12\)}}

\put(92,10){\circle*{0.9}}

\put(88,15){\makebox(0,0)[bl]{Worst}}
\put(88,12){\makebox(0,0)[bl]{point}}


\end{picture}
\end{center}

  {\bf Example 2.}
 Ordinal estimates of DAs are integrated into the resultant ordinal estimates for modular solutions
 (via method of integration tables, Fig. 21):~
 \(\{ e(T_{1}) = 1 \}\),
 \(\{ e(T_{2}) = 4 \}\),
 \(\{ e(T_{3}) = 3 \}\), and
 \(\{ e(T_{4}) = 3 \}\).

\begin{center}
\begin{picture}(75,112)
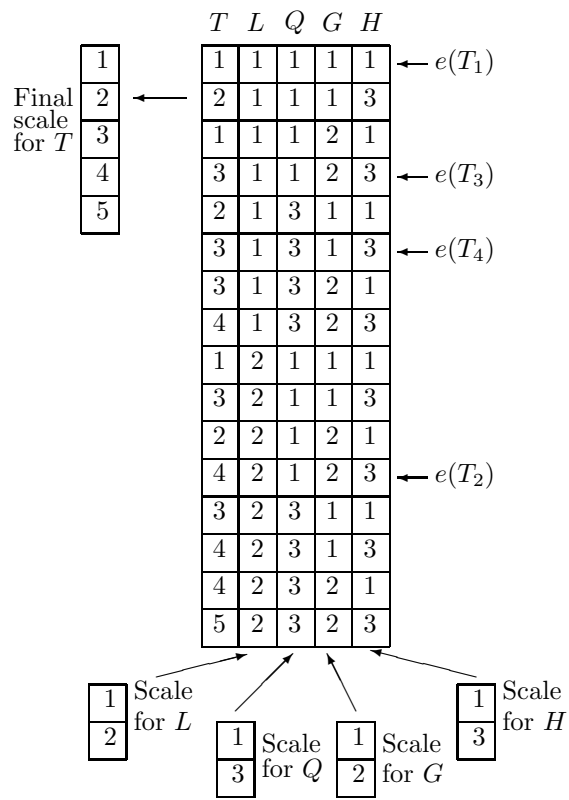
\put(00,00){\makebox(0,0)[bl]{Fig. 21. Integration of by
 tables for example 2}}


\put(09,80){\line(0,1){25}} \put(14,80){\line(0,1){25}}
\put(09,80){\line(1,0){05}}

\put(09,85){\line(1,0){05}} \put(09,90){\line(1,0){05}}
\put(09,95){\line(1,0){05}} \put(09,100){\line(1,0){05}}
\put(09,105){\line(1,0){05}}

\put(11,102){\makebox(0,8)[bl]{\(1\)}}
\put(11,97){\makebox(0,8)[bl]{\(2\)}}
\put(11,92){\makebox(0,8)[bl]{\(3\)}}
\put(11,87){\makebox(0,8)[bl]{\(4\)}}
\put(11,82){\makebox(0,8)[bl]{\(5\)}}

\put(00,97){\makebox(0,8)[bl]{Final}}
\put(00,94){\makebox(0,8)[bl]{scale}}
\put(00,91){\makebox(0,8)[bl]{for \(T\)}}




\put(10,10){\line(0,1){10}} \put(15,10){\line(0,1){10}}

\put(10,10){\line(1,0){05}} \put(10,15){\line(1,0){05}}
\put(10,20){\line(1,0){05}}

\put(12,17){\makebox(0,8)[bl]{\(1\)}}
\put(12,12){\makebox(0,8)[bl]{\(2\)}}


\put(16,18){\makebox(0,8)[bl]{Scale}}
\put(16,14){\makebox(0,8)[bl]{for \(L\)}}

\put(19,21){\vector(4,1){12}}


\put(27,05){\line(0,1){10}} \put(32,05){\line(0,1){10}}
\put(27,05){\line(1,0){05}} \put(27,10){\line(1,0){05}}
\put(27,15){\line(1,0){05}}


\put(29,12){\makebox(0,8)[bl]{\(1\)}}
\put(29,07){\makebox(0,8)[bl]{\(3\)}}

\put(33,11){\makebox(0,8)[bl]{Scale}}
\put(33,07){\makebox(0,8)[bl]{for \(Q\)}}

\put(29.5,16.5){\vector(1,1){07.6}}


\put(43,05){\line(0,1){10}} \put(48,05){\line(0,1){10}}

\put(43,05){\line(1,0){05}} \put(43,10){\line(1,0){05}}
\put(43,15){\line(1,0){05}}


\put(45,12){\makebox(0,8)[bl]{\(1\)}}
\put(45,07){\makebox(0,8)[bl]{\(2\)}}

\put(45.5,16.5){\vector(-1,2){3.8}}


\put(49,11){\makebox(0,8)[bl]{Scale}}
\put(49,07){\makebox(0,8)[bl]{for \(G\)}}


\put(61.5,21){\vector(-4,1){14}}

\put(59,10){\line(0,1){10}} \put(64,10){\line(0,1){10}}


\put(59,10){\line(1,0){05}} \put(59,15){\line(1,0){05}}
\put(59,20){\line(1,0){05}}

\put(61,17){\makebox(0,8)[bl]{\(1\)}}
\put(61,12){\makebox(0,8)[bl]{\(3\)}}


\put(65,18){\makebox(0,8)[bl]{Scale}}
\put(65,14){\makebox(0,8)[bl]{for \(H\)}}


\put(23,98){\vector(-1,0){7}}

\put(25,25){\line(0,1){80}} \put(30,25){\line(0,1){80}}
\put(35,25){\line(0,1){80}} \put(40,25){\line(0,1){80}}
\put(45,25){\line(0,1){80}} \put(50,25){\line(0,1){80}}

\put(29.8,25){\line(0,1){80}} \put(29.9,25){\line(0,1){80}}

\put(25,25){\line(1,0){25}} \put(25,30){\line(1,0){25}}
\put(25,35){\line(1,0){25}} \put(25,40){\line(1,0){25}}
\put(25,45){\line(1,0){25}} \put(25,50){\line(1,0){25}}
\put(25,55){\line(1,0){25}} \put(25,60){\line(1,0){25}}
\put(25,65){\line(1,0){25}} \put(25,70){\line(1,0){25}}
\put(25,75){\line(1,0){25}} \put(25,80){\line(1,0){25}}
\put(25,85){\line(1,0){25}}

\put(25,90){\line(1,0){25}} \put(25,95){\line(1,0){25}}
\put(25,100){\line(1,0){25}} \put(25,105){\line(1,0){25}}


\put(26,107){\makebox(0,8)[bl]{\(T\)}}
\put(31,107){\makebox(0,8)[bl]{\(L\)}}
\put(36,106.5){\makebox(0,8)[bl]{\(Q\)}}
\put(41,107){\makebox(0,8)[bl]{\(G\)}}
\put(46,107){\makebox(0,8)[bl]{\(H\)}}




\put(56,101){\makebox(0,8)[bl]{\(e(T_{1})\)}}
\put(55,102.5){\vector(-1,0){4}}


\put(56,46){\makebox(0,8)[bl]{\(e(T_{2})\)}}
\put(55,47.5){\vector(-1,0){4}}


\put(56,86){\makebox(0,8)[bl]{\(e(T_{3})\)}}
\put(55,87.5){\vector(-1,0){4}}


\put(56,76){\makebox(0,8)[bl]{\(e(T_{4})\)}}
\put(55,77.5){\vector(-1,0){4}}



\put(26.5,102){\makebox(0,8)[bl]{\(1\)}}
\put(26.5,97){\makebox(0,8)[bl]{\(2\)}}
\put(26.5,92){\makebox(0,8)[bl]{\(1\)}}
\put(26.5,87){\makebox(0,8)[bl]{\(3\)}}

\put(26.5,82){\makebox(0,8)[bl]{\(2\)}}
\put(26.5,77){\makebox(0,8)[bl]{\(3\)}}
\put(26.5,72){\makebox(0,8)[bl]{\(3\)}}
\put(26.5,67){\makebox(0,8)[bl]{\(4\)}}

\put(26.5,62){\makebox(0,8)[bl]{\(1\)}}
\put(26.5,57){\makebox(0,8)[bl]{\(3\)}}

\put(26.5,52){\makebox(0,8)[bl]{\(2\)}}
\put(26.5,47){\makebox(0,8)[bl]{\(4\)}}


\put(26.5,42){\makebox(0,8)[bl]{\(3\)}}
\put(26.5,37){\makebox(0,8)[bl]{\(4\)}}
\put(26.5,32){\makebox(0,8)[bl]{\(4\)}}
\put(26.5,27){\makebox(0,8)[bl]{\(5\)}}


\put(31.5,102){\makebox(0,8)[bl]{\(1\)}}
\put(31.5,97){\makebox(0,8)[bl]{\(1\)}}
\put(31.5,92){\makebox(0,8)[bl]{\(1\)}}
\put(31.5,87){\makebox(0,8)[bl]{\(1\)}}

\put(31.5,82){\makebox(0,8)[bl]{\(1\)}}
\put(31.5,77){\makebox(0,8)[bl]{\(1\)}}
\put(31.5,72){\makebox(0,8)[bl]{\(1\)}}
\put(31.5,67){\makebox(0,8)[bl]{\(1\)}}

\put(31.5,62){\makebox(0,8)[bl]{\(2\)}}
\put(31.5,57){\makebox(0,8)[bl]{\(2\)}}
\put(31.5,52){\makebox(0,8)[bl]{\(2\)}}
\put(31.5,47){\makebox(0,8)[bl]{\(2\)}}

\put(31.5,42){\makebox(0,8)[bl]{\(2\)}}
\put(31.5,37){\makebox(0,8)[bl]{\(2\)}}
\put(31.5,32){\makebox(0,8)[bl]{\(2\)}}
\put(31.5,27){\makebox(0,8)[bl]{\(2\)}}


\put(36.5,102){\makebox(0,8)[bl]{\(1\)}}
\put(36.5,97){\makebox(0,8)[bl]{\(1\)}}
\put(36.5,92){\makebox(0,8)[bl]{\(1\)}}
\put(36.5,87){\makebox(0,8)[bl]{\(1\)}}

\put(36.5,82){\makebox(0,8)[bl]{\(3\)}}
\put(36.5,77){\makebox(0,8)[bl]{\(3\)}}
\put(36.5,72){\makebox(0,8)[bl]{\(3\)}}
\put(36.5,67){\makebox(0,8)[bl]{\(3\)}}

\put(36.5,62){\makebox(0,8)[bl]{\(1\)}}
\put(36.5,57){\makebox(0,8)[bl]{\(1\)}}
\put(36.5,52){\makebox(0,8)[bl]{\(1\)}}
\put(36.5,47){\makebox(0,8)[bl]{\(1\)}}

\put(36.5,42){\makebox(0,8)[bl]{\(3\)}}
\put(36.5,37){\makebox(0,8)[bl]{\(3\)}}
\put(36.5,32){\makebox(0,8)[bl]{\(3\)}}
\put(36.5,27){\makebox(0,8)[bl]{\(3\)}}


\put(41.5,102){\makebox(0,8)[bl]{\(1\)}}
\put(41.5,97){\makebox(0,8)[bl]{\(1\)}}
\put(41.5,92){\makebox(0,8)[bl]{\(2\)}}
\put(41.5,87){\makebox(0,8)[bl]{\(2\)}}

\put(41.5,82){\makebox(0,8)[bl]{\(1\)}}
\put(41.5,77){\makebox(0,8)[bl]{\(1\)}}
\put(41.5,72){\makebox(0,8)[bl]{\(2\)}}
\put(41.5,67){\makebox(0,8)[bl]{\(2\)}}

\put(41.5,62){\makebox(0,8)[bl]{\(1\)}}
\put(41.5,57){\makebox(0,8)[bl]{\(1\)}}
\put(41.5,52){\makebox(0,8)[bl]{\(2\)}}
\put(41.5,47){\makebox(0,8)[bl]{\(2\)}}

\put(41.5,42){\makebox(0,8)[bl]{\(1\)}}
\put(41.5,37){\makebox(0,8)[bl]{\(1\)}}
\put(41.5,32){\makebox(0,8)[bl]{\(2\)}}
\put(41.5,27){\makebox(0,8)[bl]{\(2\)}}


\put(46.5,102){\makebox(0,8)[bl]{\(1\)}}
\put(46.5,97){\makebox(0,8)[bl]{\(3\)}}
\put(46.5,92){\makebox(0,8)[bl]{\(1\)}}
\put(46.5,87){\makebox(0,8)[bl]{\(3\)}}

\put(46.5,82){\makebox(0,8)[bl]{\(1\)}}
\put(46.5,77){\makebox(0,8)[bl]{\(3\)}}
\put(46.5,72){\makebox(0,8)[bl]{\(1\)}}
\put(46.5,67){\makebox(0,8)[bl]{\(3\)}}

\put(46.5,62){\makebox(0,8)[bl]{\(1\)}}
\put(46.5,57){\makebox(0,8)[bl]{\(3\)}}
\put(46.5,52){\makebox(0,8)[bl]{\(1\)}}
\put(46.5,47){\makebox(0,8)[bl]{\(3\)}}

\put(46.5,42){\makebox(0,8)[bl]{\(1\)}}
\put(46.5,37){\makebox(0,8)[bl]{\(3\)}}
\put(46.5,32){\makebox(0,8)[bl]{\(1\)}}
\put(46.5,27){\makebox(0,8)[bl]{\(3\)}}

\end{picture}
\end{center}

 {\bf Example 3.} Vector-like (two-element) estimates are
 integrated into an ordinal scale for modular solutions:
 (1) summarization  (by vector-estimate components) for each modular solution
 (i.e., \(T_{1}\),\(T_{2}\),\(T_{3}\),\(T_{4}\)),
 (2) selection of Pareto-efficient solutions) (Fig. 22):~

 (a) vector-like estimates:~
  \(e(T_{1}) = (5,5) \),
  \(e(T_{2}) = (9,8) \),
 \(e(T_{3}) = (9,7) \),  and
 \(e(T_{4}) = (8,8) \);

 (b) domination (preferences):~
 \(T_{1}  \succ  T_{2}\),
 \(T_{1} \succ T_{3}\),
 \(T_{1} \succ T_{4}\),
 \(T_{3} \succ T_{2}\),
 and
 \(T_{4} \succ T_{2}\).

 (c) the resultant ordinal scale (type \(D\)):~
 the layer of Pareto-efficient solution (layer \(1\)):~
 \(\{ T_{1} \}\);
 the next layer (layer \(2\)):~
 \(\{ T_{3}, T_{4} \}\);
 the next layer (layer \(3\)):~
 \(\{T_{2} \}\).

 Thus, the resultant priorities are obtained:~
 \(r(T_{1}) =1\),
 \(r(T_{2}) =3\),
 \(r(T_{3}) =2\), and
 \(r(T_{4}) =2\).

 {\bf Example 4.}
 Ordinal estimates of DAs are transformed into poset-like  estimate for modular solutions
 (Fig. 23),
 selection of Pareto-efficient solutions:

 (a) poset-like estimates:~
  \(n(T_{1}) = (4,0,0) \),
  \(n(T_{2}) = (1,2,1) \),
 \(n(T_{3}) = (2,1,1) \),  and
 \(n(T_{4}) = (2,1,1) \);

 (b) domination (preferences):~
 \(T_{1} \succ T_{2}\),
 \(T_{1} \succ T_{3}\),
 \(T_{1} \succ T_{4}\),
 \(T_{3} \succ T_{2}\),
 \(T_{4} \succ T_{2}\);

 (c) the resultant ordinal scale (type \(D\)):~
 the layer of Paret-efficient solutions (layer \(1\)):~
 \(\{ T_{1}\}\),
 the next layer (layer \(2\)):~
 \(\{ T_{3}, T_{4} \}\),
 the next layer (layer \(3\)):~
 \(\{ T_{2}\}\).

 Thus, the resultant priorities are obtained:~
 \(r(T_{1}) =1\),
 \(r(T_{2}) =3\),
 \(r(T_{3}) =2\), and
 \(r(T_{4}) =2\).

 .

\begin{center}
\begin{picture}(63,55)

\put(01,00){\makebox(0,0)[bl]{Fig. 22. Multicriteria
 description}}


\put(19,23){\circle*{0.7}} \put(19,23){\circle{2.1}}
\put(10.5,22){\makebox(0,0)[bl]{Best}}
\put(10.5,19){\makebox(0,0)[bl]{point}}

\put(00,08.5){\makebox(0,0)[bl]{\((0,0)\)}}

\put(04,13){\vector(0,1){35}} \put(04,13){\vector(1,0){49}}


\put(19,11.5){\line(0,1){3}} \put(34,11.5){\line(0,1){3}}
\put(49,11.5){\line(0,1){3}}

\put(33,08){\makebox(0,0)[bl]{\(8\)}}
\put(18,08){\makebox(0,0)[bl]{\(4\)}}


\put(23,25){\circle*{1}} \put(23,25){\circle{1.8}}
\put(24.4,23.5){\makebox(0,0)[bl]{\(T_{1}\)}}


\put(34,37){\circle*{1}} \put(34,37){\circle{1.8}}
\put(35,35.5){\makebox(0,0)[bl]{\(T_{2}\)}}


\put(30,37){\circle*{1}} \put(30,37){\circle{1.8}}
\put(25.5,35.5){\makebox(0,0)[bl]{\(T_{3}\)}}


\put(34,33){\circle*{1}} \put(34,33){\circle{1.8}}
\put(35,31.5){\makebox(0,0)[bl]{\(T_{4}\)}}


\put(02.5,23){\line(1,0){3}} \put(02.5,33){\line(1,0){3}}
\put(02.5,43){\line(1,0){3}}

\put(00,39.7){\makebox(0,0)[bl]{\(12\)}}
\put(00,32){\makebox(0,0)[bl]{\(8\)}}
\put(00,22){\makebox(0,0)[bl]{\(4\)}}


\put(38,09){\makebox(0,0)[bl]{Criterion 2}}
\put(37,05){\makebox(0,0)[bl]{(``Physics'')}}

\put(00,52){\makebox(0,0)[bl]{Criterion 1}}
\put(00,48){\makebox(0,0)[bl]{(``Mathematics'')}}


\put(07,43){\line(1,0){4}} \put(13,43){\line(1,0){4}}
\put(19,43){\line(1,0){4}} \put(25,43){\line(1,0){4}}
\put(31,43){\line(1,0){4}} \put(37,43){\line(1,0){4}}
\put(43,43){\line(1,0){4}}



\put(49,16){\line(0,1){4}} \put(49,22){\line(0,1){4}}
\put(49,28){\line(0,1){4}}

\put(49,34){\line(0,1){4}} \put(49,40){\line(0,1){4}}

\put(49,43){\circle*{1.8}}

\put(43,52){\makebox(0,0)[bl]{Worst}}
\put(43,49){\makebox(0,0)[bl]{point}}
\put(42,45){\makebox(0,0)[bl]{\((12,12)\)}}

\end{picture}
\begin{picture}(54,107)

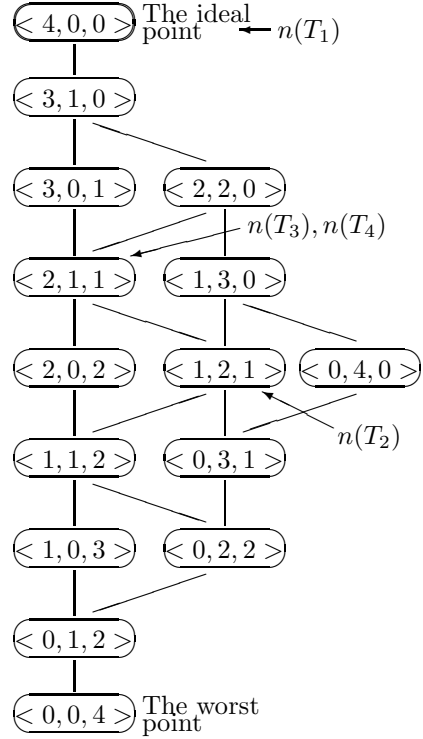
\put(00,00){\makebox(0,0)[bl] {Fig. 23. Poset
 \(n(T)=(\eta_{1},\eta_{2},\eta_{3})\)}}


\put(17,101){\makebox(0,0)[bl]{The ideal}}
\put(17,98){\makebox(0,0)[bl]{point}}

\put(00,99){\makebox(0,0)[bl]{\(<4,0,0>\) }}
\put(08,101){\oval(16,5)} \put(08,101){\oval(15.7,4.7)}

\put(08,94){\line(0,1){4}}

\put(00,89){\makebox(0,0)[bl]{\(<3,1,0>\)}}
\put(08,91){\oval(16,5)}
\put(08,82){\line(0,1){6}}

\put(00,77){\makebox(0,0)[bl]{\(<3,0,1>\) }}
\put(08,79){\oval(16,5)}



\put(08,70){\line(0,1){6}}
\put(00,65){\makebox(0,0)[bl]{\(<2,1,1>\)}}
\put(08,67){\oval(16,5)}


\put(08,58){\line(0,1){6}}
\put(00,53){\makebox(0,0)[bl]{\(<2,0,2>\) }}
\put(08,55){\oval(16,5)}

%
\put(35,98){\makebox(0,0)[bl]{\(n(T_{1})\)}}
\put(34,100){\vector(-1,0){04}}

%
\put(31,72){\makebox(0,0)[bl]{\(n(T_{3}),n(T_{4})\)}}

\put(30,73.5){\vector(-4,-1){14.5}}

%
\put(43,44){\makebox(0,0)[bl]{\(n(T_{2})\)}}

\put(42.5,47){\vector(-2,1){09.5}}


\put(08,46){\line(0,1){6}}
\put(00,41){\makebox(0,0)[bl]{\(<1,1,2>\) }}
\put(08,43){\oval(16,5)}

\put(08,34){\line(0,1){6}}
\put(00,29){\makebox(0,0)[bl]{\(<1,0,3>\) }}
\put(08,31){\oval(16,5)}

\put(08,22){\line(0,1){6}}
\put(00,17){\makebox(0,0)[bl]{\(<0,1,2>\) }}
\put(08,19){\oval(16,5)}

\put(08,12){\line(0,1){4}}

\put(00,07){\makebox(0,0)[bl]{\(<0,0,4>\) }}
\put(08,09){\oval(16,5)}

\put(17,09){\makebox(0,0)[bl]{The worst}}
\put(17,06){\makebox(0,0)[bl]{point}}


\put(25.5,82.5){\line(-3,1){15}}

\put(25.5,75.5){\line(-3,-1){15}}

\put(20,77){\makebox(0,0)[bl]{\(<2,2,0>\) }}
\put(28,79){\oval(16,5)}

\put(28,70){\line(0,1){6}}
\put(20,65){\makebox(0,0)[bl]{\(<1,3,0>\) }}
\put(28,67){\oval(16,5)}

\put(10.5,63.5){\line(3,-1){15}}

\put(28,58){\line(0,1){6}}
\put(20,53){\makebox(0,0)[bl]{\(<1,2,1>\) }}
\put(28,55){\oval(16,5)}
\put(10.5,46.5){\line(3,1){15}}

\put(28,46){\line(0,1){6}}

\put(20,41){\makebox(0,0)[bl]{\(<0,3,1>\) }}
\put(28,43){\oval(16,5)}
\put(10.5,39.5){\line(3,-1){15}}

\put(28,34){\line(0,1){6}}
\put(20,29){\makebox(0,0)[bl]{\(<0,2,2>\) }}
\put(28,31){\oval(16,5)}
\put(25.5,27.5){\line(-3,-1){15}}


\put(45.5,58.5){\line(-3,1){15}} \put(45.5,51.5){\line(-3,-1){15}}

\put(38,53){\makebox(0,0)[bl]{\(<0,4,0>\) }}
\put(46,55){\oval(16,5)}

\end{picture}
\end{center}

 {\bf Example 5.}
  Interval multiset estimates of DAs (poset-like scale, Fig. 7)
  are transformed
 (searching for the median-like estimate)
   into  interval multiset estimates
  for modular solutions
 (poset-like scale, Fig. 7),
 selection of Pareto-efficient solutions:

 (a) interval multiset estimates:~
  \(n(T_{1}) = (3,1,0) \),
  \(n(T_{2}) = (0,4,0) \),
 \(n(T_{3}) = (1,3,0) \),  and
 \(n(T_{4}) = (1,3,0) \);

 (b) domination (preferences):~
 \(T_{1} \succ T_{2}\),
 \(T_{1} \succ T_{3}\),
 \(T_{1} \succ T_{4}\),
 \(T_{3} \succ T_{2}\),
 \(T_{4} \succ T_{2}\);

 (c) the resultant ordinal scale (type \(D\)):~
  the layer of Paret-efficient solution (layer \(1\)):~
 \(\{ T_{1}\}\),
 the next layer (layer \(2\)):~
 \(\{ T_{3}, T_{4} \}\),
 the next layer (layer \(3\)):~
 \(\{ T_{2} \}\).

 Thus, the resultant priorities are obtained:~
 \(r(T_{1}) =1\),
 \(r(T_{2}) =3\),
 \(r(T_{3}) =2\), and
 \(r(T_{4}) =2\).

\section{Conclusion}

 This survey paper briefly
   described approaches to
 evaluation
 of composite (modular) systems.
%
%
%
 In the future, it may be  reasonable to consider
 the following research directions:
 (1) study of other scale transformation problems
 (e.g., ~{\it poset} \(\lambda\) \(\Rightarrow\) {\it poset} \(\mu\) ),
 (2) study of multi-stage
 scale transformation procedures (frameworks),
 (3) examination of various  real-world applications,
 (e.g., usage of stochastic models, fuzzy sets),
 (4) analysis and usage of reference solutions;
 (5) taking into account uncertainty,
 (6) special analysis of the correspondence between
 considered system evaluation problems,  scale
 transformation problems, and traditional decision making problems,
 (7) additional attention to issues of system component
 compatibility assessment and integration of the corresponding
 estimates into the total system estimates,
 (8) design of a special software tool for scale transformation/integration
 (e.g., library of various scales, visualization support,
 automatic and interactive  procedures),
 and
 (9) usage of the described system evaluation approaches in
 education (computer science, engineering, applied mathematics, management).


\begin{thebibliography}{140}

 \bibitem {alek76} V.B. Alekseev,
 Deciphering algorithm of some classes of monotonic many-valued
 functions.
  {\it USSR Computational Mathematics and Mathematical Physics},
  16(1) (1976) 180--189.

  \bibitem {bald00} C.Y. Baldwin, K.B. Clark,
  {\it Design Rules: The Power of Modularity}.
   MIT Press, Cambridge, Mass., 2000.

  \bibitem {borg05} I. Borg, P. Groenen,
 {\it Modern Multidimensional Scaling: theory and applications}.
 2nd ed., Springer, New York, 2005.

  \bibitem {brans84} J.P. Brans, B. Mareschal, Ph. Vincke,
 PROMETHEE: A new family of outranking methods in multicriteria
 analysis.
 In: Brans J.P. (Ed)
 {\it Operations Research '84}, North-Holland, New York,
 pp. 477--490, 1984.

 \bibitem {clune12} J. Clune, J.-B. Mouret, H. Lipson,
 The evolutionary origins of modularity.
  Electronic preprint, 17 pp.,
 July 11, 2012.
 http://arxiv.org/abs/1207.2743 [q-bio.PE]

  \bibitem {fis70} P.C. Fishburn,
 {\it Utility Theory for Decision Making,}
  J.Wiley\& Sons, New York, 1970.

  \bibitem {glo84} V.A. Glotov, V.V. Paveljev,
  {\it Vector Stratification}.  Nauka, Moscow, 1984 (in Russian).

  \bibitem {hua98} C.C. Huang, A. Kusiak,
 Modularity in design of products and systems.
 {\it  IEEE Trans. SMC - Part A},  28(10) (1998) 66--77.

  \bibitem {kee76} R.L. Keeny, H. Raiffa,
 {\it Decisions with Multiple Objectives: Preferences and Value
 Tradeoffs}.
  J.Wiley\& Sons, New York, 1976.

  \bibitem {la94} Y.-J. Lai, T.-Y. Liu, C.-L. Hwang,
 TOPSIS for MODM.
 {\it EJOR}, 76(3) (1994) 486--500.

 \bibitem {larmf86} O.I. Larichev, H.M. Moshkovich, E.M. Furems,
 Decision support system 'Class'.
 In: B. Brehmer, H. Jungerman, P. Lourens, G. Sevon (Eds.),
 New Direction in Research on Decision Making. North Holland,
 Amsterdam, Elsevier, 305--315, 1988.

 \bibitem {lev98} M.Sh. Levin,
  {\it Combinatorial Engineering of Decomposbale Systems}.
   Kluwer,
   Boston, 1998.

  \bibitem {levf01} M.Sh. Levin,
  System synthesis with morphological clique problem: fusion of
  subsystem evaluation decisions.
  {\it Information Fusion} 2(3) (2001) 225--237.

 \bibitem {lev06} M.Sh. Levin,
  {\it Composite Systems Decisions}. Springer, New York, 2006.

  \bibitem {lev09} M.Sh. Levin,
 Combinatorial optimization in system configuration design.
  {\it Automation and Remote Control}  70(3) (2009) 519--561.

  \bibitem {lev11agg} M.Sh. Levin,
 Aggregation of composite solutions: strategies, models,
 examples.
 Electronic preprint. 72 pp., Nov. 29, 2011.
 http://arxiv.org/abs/1111.6983 [cs.SE]

  \bibitem{lev12morph} M.Sh. Levin,
 Morphological methods for design of modular systems (a survey).
  Electronic  preprint. 20 pp., Jan. 9, 2012.
  http://arxiv.org/abs/1201.1712 [cs.SE]

  \bibitem{lev12a} M.Sh. Levin,
   Multiset estimates and combinatorial synthesis.
    Electronic preprint. 30 pp., May 9, 2012.
    http://arxiv.org/abs/1205.2046 [cs.SY]

  \bibitem {lev12b} M.Sh. Levin,
  Composite strategy for multicriteria ranking/sorting
 (methodological issues, examples).
    Electronic preprint. 24 pp., Nov. 9, 2012.
    http://arxiv.org/abs/1211.2245 [math.OC]

    \bibitem {lev13intro} M.Sh. Levin,
  Note on combinatorial engineering frameworks
 for hierarchical modular systems.
    Electronic preprint. 11 pp., Mar. 29, 2013.
    http://arxiv.org/abs/1304.0030 [math.OC]

  \bibitem {levmih88} M.Sh. Levin, A.A. Michailov,
  {\it Fragments of Objects Set Stratification Technology},
  Preprint, Moscow, Inst. for System Analysis, Moscow, 1988
  (in Russian).

  \bibitem {mirkin79}  B.G. Mirkin,
 {\it Group Choice}. Halsted Press, 1979.

 \bibitem {pareto71} V. Pareto,
 {\it Mannual of Political Economy}.
 (English translation), A.M. Kelley Publ.,
 New York, 1971.

  \bibitem {roy96} B. Roy,
 {\it Multicriteria Methodology for Decision Aiding}.
  Kluwer Academic Publishers, Dordrecht, 1996.

  \bibitem {saaty88} T.L. Saaty,
  {\it The Analytic Hierarchy Process}.
   MacGraw-Hill, New York, 1988.

 \bibitem {ser83} A.V. Serzhantov,
 An optimal deciphering algorithm for some classes of monotonic functions.
  {\it USSR Computational Mathematics and Mathematical Physics},
   23(1) (1983)  144--148.

 \bibitem {shih07} H.-S. Shih, H.-J. Shyur, E.S. Lee,
 An extension of TOPSIS for group decision making.
 {\it Mathematical and Computer Modelling}
  45(7-8) (2007) 801--813.

 \bibitem {ste86} R.E. Steuer,
  {\it Multple Criteria Optimization: Theory, Computation, and Application}.
   J.Wiley\& Sons, New York, 1986.

 \bibitem {than01} E. Thanassoulis,
 {\it Introduction to the Theory and Application of Data Envelopment Analysis:
 A foundation text with integrated software}.
  Kluwer Academic Publishers, Boston, 2001.

  \bibitem {zap02} C. Zopounidis, M. Doumpos,
  Multicriteria classification and sorting methods:
  a literature review. {\it EJOR} 138(2) (2002) 229–-246.



\end{thebibliography}
\end{document}